%% file: main.tex
\documentclass{article}

\usepackage[main, final]{neurips_2026}

\usepackage[utf8]{inputenc} %
\usepackage[T1]{fontenc}    %
\usepackage{hyperref}       %
\usepackage{url}            %
\usepackage{booktabs}       %
\usepackage{amsfonts, amssymb}       %
\usepackage{nicefrac}       %
\usepackage{microtype}      %
\usepackage{xcolor}         %
\usepackage{bm}
\usepackage{amsmath}
\usepackage{amsthm}

\usepackage{algorithm}
\usepackage{algpseudocode}
\usepackage{graphicx}
\usepackage{subcaption}
\usepackage{enumitem}
\usepackage{makecell}

\newtheorem{remark}{Remark}
\newtheorem{theorem}{Theorem}
\newtheorem{mylemma}[theorem]{Lemma}
\newtheorem{lemma}[theorem]{Lemma}
\newtheorem{proposition}[theorem]{Proposition}

  \DeclareMathOperator{\tr}{tr}
  \DeclareMathOperator{\Haar}{Haar}
  \DeclareMathOperator{\Unif}{Unif}
  \DeclareMathOperator{\Cov}{Cov}
   \DeclareMathOperator{\Var}{Var}
   \DeclareMathOperator{\Dir}{Dir}

  \newcommand{\E}{\mathbb{E}}
  \newcommand{\1}{\mathbf{1}}

\usepackage{tcolorbox}

\DeclareMathOperator*{\argmin}{arg\,min}

\title{A Transformer-Derived Iterative Preconditioner}

\author{%
  Patrick Lutz\thanks{corresponding author \texttt{plutz@bu.edu}}\quad Themistoklis Haris\quad Aditya Gangrade\quad Venkatesh Saligrama \\
    Departments of Computer Science and Electrical and Computer Engineering\\
  Boston University\\
  \texttt{\{plutz,tharis,gangrade,srv\}@bu.edu} \\
}

\begin{document}

\maketitle

\begin{abstract}
  \input{paper/abstract3}
\end{abstract}

\input{paper/introduction_v4}
\input{paper/related_v4}
\input{paper/tools_rev4}

\input{paper/emergence2}

\input{paper/practical2}

\input{paper/simulations2}

\input{paper/analysis3}

\input{paper/conclusion3}

\bibliographystyle{plainnat}
\bibliography{paper/bibliography}

\newpage
\appendix

\input{paper/expts_appx}

\input{paper/simulations_appendix2}

\input{paper/analysis_appx4}



\end{document}

%% file: paper/abstract3.tex
Transformers trained on in-context prediction problems can learn known numerical algorithms, e.g., gradient descent. We go further, showing that their internal computations can reveal \textit{new} algorithms with provable guarantees. A softmax-attention transformer trained to solve linear systems $Xw=y$ implicitly improves matrix conditioning. Tracing $XX^\top$ reveals an iterative preconditioning primitive, which we develop into RAPID, a standalone algorithm requiring no learned components.

RAPID is an anytime preconditioner for symmetric positive definite (SPD) matrices, using randomized mixing and sparse local corrections to improve conditioning. Proof-of-concept experiments on synthetic and real-world problems consistently improve conditioning and reduce conjugate gradient iterations needed to reach a target accuracy. For a Haar-mixing idealization, we prove convergence to any fixed target condition number regardless of spectral distribution, with only logarithmic dependence on initial conditioning. Together, these results show how a transformer can discover a broadly applicable numerical algorithm.

%% file: paper/introduction_v4.tex
\section{Introduction}

Transformers can learn from examples in their input at test time without updating their parameters, a capability known as in-context learning \citep{brown2020language,garg22can}. Studies of in-context linear regression have shown that transformers can implement familiar numerical algorithms, including gradient descent, preconditioned gradient descent, and Newton-like methods \citep{akyurek23learning,vonoswald23transformers,ahn23transfromers,fu2024transformers}. We ask:
\begin{center}
    \textit{Can studying learned transformer computations lead to new, general-purpose numerical algorithms that work independently of the transformer and beyond its training distribution?}
\end{center}
We answer affirmatively by uncovering a new preconditioning primitive in a trained transformer's computations and developing it into a standalone numerical algorithm.

\textbf{The resulting algorithm: RAPID.}
Preconditioning is often crucial for efficiently solving symmetric positive definite (SPD) systems $Ax=b$. Conjugate gradient (CG) avoids direct factorization, but its convergence depends strongly on the spectrum of $A$ \citep{hestenes1952methods,saad2003iterative}. Preconditioners improve convergence by transforming the system into an equivalent, better-conditioned one \citep{benzi2002preconditioning,wathen2015preconditioning}. Classical approaches often construct approximate factorizations, inverses, or surrogate operators \citep{meijerink1981guidelines,benzi1996sparse}.

We introduce RAPID (Randomized Anytime Preconditioning by Iterative Dynamics), which improves global conditioning through simple local corrections. Its key idea is to repeatedly randomize the coordinate system, exposing different matrix interactions that can be corrected locally. Specifically, RAPID alternates two steps. First, random orthogonal mixing redistributes matrix interactions without changing the spectrum. Second, each row selects its smallest off-diagonal entry and applies a sparse correction that acts to weaken the selected coupling. Repeating these steps allows local corrections to progressively improve global conditioning without explicitly computing eigenvectors or matrix inverses.

RAPID accumulates these transformations into a preconditioning factor $P_t$, yielding $A_t=P_tAP_t^\top$. Each iteration produces a valid preconditioner, making RAPID an anytime algorithm. Fast structured mixing and sparse corrections enable efficient, highly parallel dense implementations. The underlying update is derived from the trained transformer's internal computation, as described next.

\textbf{Algorithm discovery.}
We train a softmax-attention transformer on in-context linear systems $Xw=y$, where $X$ need not be SPD. Layerwise orthogonal symmetry interventions \citep{lutz2026layerwise} allow us to trace the evolution of the Gram matrix $G_\ell=X_\ell X_\ell^\top$ during prediction. We observe that its conditioning improves and identify a simple learned congruence update responsible for this effect. Crucially, this update depends only on $G_\ell$, allowing RAPID to operate directly on arbitrary SPD matrices without the cost of forming a Gram matrix.

The connection to RAPID follows naturally: the symmetry intervention introduces random rotations, while learned attention applies local corrections that tend to weaken strong matrix interactions. The residual update applies the same row transformation to examples and labels, inducing the congruence update. To obtain an efficient algorithm, RAPID takes the low-temperature limit of softmax to select sparse off-diagonal row-min corrections and replaces dense rotations with fast structured mixing \citep{yu2016orthogonal}. The resulting algorithm requires no learned components.

\textbf{Results and guarantees.}
We evaluate RAPID on four synthetic spectral families and $31$
SuiteSparse SPD matrices, well beyond the transformer's
training distribution. RAPID substantially improves conditioning
across all four families and, remarkably, on every SuiteSparse
matrix at every tested step count. In contrast, randomized
Nystr\"om preconditioning is more sensitive to spectral structure.
These improvements also translate into substantially fewer
conjugate gradient (CG) iterations to reach a fixed accuracy.
Progressive gains with additional RAPID steps illustrate
its anytime nature.

Theoretically, we prove that an idealized version of RAPID, using Haar mixing and block-restricted row-min corrections, reaches any fixed target condition number in $ O(d\log d \log\kappa(A))$ expected iterations. This holds for any SPD matrix $A$, regardless of its eigenvalue distribution. At this iteration scale, RAPID's per-step cost of $O(d^2\log d)$ corresponds to near-cubic setup complexity, comparable to classical dense factorization. Unlike direct factorization, however, RAPID can stop early to balance setup cost against preconditioner quality. Although the iteration-complexity guarantee remains limited to an idealized variant, experiments show that its assumptions do not accelerate conditioning on tested spectra, suggesting they facilitate the proof rather than the algorithm's effectiveness.

\textbf{Contributions.}
We make three core contributions:\vspace{-.4\baselineskip}
\begin{itemize}[nosep,leftmargin=*]
    \item We discover a new preconditioning primitive in a trained transformer: a solution-preserving update to the linear system $Xw=y$ that improves the conditioning of the Gram matrix $XX^\top$.
    \item We develop this mechanism into RAPID, a standalone anytime preconditioner for arbitrary SPD matrices, requiring no learned components and using efficient, parallelizable iterations.
    \item We establish theoretical and empirical evidence of RAPID's effectiveness: an idealized variant provably reaches any fixed target condition number for arbitrary SPD matrices, while dense experiments demonstrate broad conditioning gains and substantially faster CG convergence.
\end{itemize}

%% file: paper/related_v4.tex
\section{Related Work}

\noindent\textbf{Transformers as algorithm learners.}
For linear prediction, trained transformers approximate least-squares predictors and learn familiar methods resembling gradient descent, ridge regression, and second-order methods \citep{garg22can,akyurek23learning,vonoswald23transformers,ahn23transfromers,zhang24transformers,fu2024transformers}. In other settings, trained transformers also learn temporal-difference updates and graph-connectivity algorithms \citep{wang2025transformers,ye2026transformers}. Beyond known methods, transformers have revealed novel optimization strategies for noisy linear regression \citep{vladymyrov2024linear} and iterative methods for masked matrix completion \citep{lutz2025linear}. We extend this direction by extracting a new conditioning primitive and developing a standalone preconditioner for arbitrary SPD matrices, with theoretical guarantees, broad conditioning gains and improved CG convergence.

\noindent\textbf{Preconditioning for SPD systems.}
Conjugate gradient (CG) solves SPD systems $Ax=b$ without direct factorization, but its convergence depends strongly on the spectrum of $A$ \citep{hestenes1952methods,saad2003iterative}. Classical preconditioners seek to improve this spectrum through incomplete Cholesky factorizations \citep{meijerink1981guidelines}, sparse approximate inverses \citep{benzi1996sparse}, or algebraic multigrid hierarchies \citep{vanvek1996algebraic}. Rather than constructing such approximations, RAPID iteratively improves conditioning through randomized mixing and sparse local corrections, accumulating a factor $P_t$ with $A_t=P_tAP_t^\top$. Each iteration yields a valid SPD preconditioned system, allowing RAPID to stop at any point without requiring a complete factorization or inverse approximation.

\textbf{Randomized preconditioning and structured transforms.}
Randomized preconditioners often construct approximations of the original problem. For least squares, sketching methods use subsampled orthogonal transforms or Gaussian projections to build smaller systems for factorization \citep{avron2010blendenpik,meng2014lsrn}; for SPD systems, randomized Nystr\"om methods construct low-rank approximations for preconditioned CG \citep{frangella2023randomized}. More broadly, fast randomized Hadamard transforms provide efficient alternatives to dense random projections \citep{ailon09fast,le2013fastfood,yu2016orthogonal}. RAPID uses fast randomized Hadamard mixing, but applies it repeatedly to expose matrix interactions for sparse local corrections rather than to construct a one-shot approximation.

\noindent\textbf{Learned preconditioners and neural solvers.}
Recent work has used neural networks to accelerate iterative linear solvers by
learning algebraic multigrid components, approximate factorizations, or
preconditioners for conjugate-gradient methods
\citep{luz2020learning,li2023learning,trifonov2026learning}.
These approaches evaluate a trained network to construct preconditioners for new systems. In contrast, we use the transformer only to discover a numerical algorithm. Once extracted, RAPID operates directly on arbitrary SPD matrices through explicit randomized updates, without learned components or neural inference at solve time.

\textbf{Symmetry and equivariance in neural networks.}
When task symmetries are known, equivariant architectures provide a way to
encode them as inductive biases, restricting the learned function class and
improving generalization
\citep{cohen2016group,kondor2018generalization,finzi2021practical}. Layerwise symmetry interventions instead use random, task-preserving coordinate changes to reveal learned transformer computations \citep{lutz2026layerwise}. This technique was originally developed for in-context classification using permutation symmetries; we adapt their identification method to the orthogonal symmetries of linear systems, revealing the conditioning primitive underlying RAPID.

%% file: paper/tools_rev4.tex
\section{Algorithm discovery through in-context learning}
\label{sec:setup-extracted-preconditioner}

In this section, we use a transformer trained on in-context linear systems to discover a numerical preconditioning primitive. We train an attention-only model to predict from examples satisfying \(Xw=y\), using a layerwise symmetry intervention to make its learned computation identifiable. We extract a two-scalar update that preserves the linear system and evolves the Gram matrix \(XX^\top\) by congruence, substantially improving its conditioning. This primitive motivates RAPID, developed in Section~\ref{sec:practical} as a standalone SPD preconditioner with no learned components.

\subsection{Transformer setup and symmetry intervention}
\label{subsec:algorithm-discovery-task}

\noindent\textbf{In-context linear systems.}
Following \citet{garg22can}, we train transformers on synthetic linear systems
\(Xw=y\), where \(X\in\mathbb R^{d\times d}\) and \(w\in\mathbb R^d\), $d=15$.
Each task samples \(X\) and a query \(x_q\in\mathbb R^d\) with independent
standard Gaussian entries, and \(w\sim\mathcal N(0,I_d/d)\), then sets
\(y=Xw\). The model observes \((X,y,x_q)\), but not \(w\), and is trained
to predict \(x_q^\top w\). Since \(w\) is resampled for every task,
prediction requires solving a new instance rather than memorizing a
fixed solution.

\noindent\textbf{Transformer architecture.}
We encode each task as
\[
    Z_0 =
    \begin{bmatrix}
        X & y \\
        x_q^\top & 0
    \end{bmatrix},
\]
with the query prediction read from the bottom-right entry of \(Z_L\) (with a negative sign) after processing through $L$ transformer layers. The attention-only transformer applies
\begin{equation}
    Z_{\ell+1}=Z_\ell+\operatorname{Attn}_{\ell}(Z_\ell),
    \qquad
    \operatorname{Attn}_{\ell}(Z)
    =
    \sigma\!\left(
        Z W_{Q,\ell}W_{K,\ell}^\top Z^\top + M
    \right)
    Z W_{V,\ell}W_{P,\ell},
    \label{eq:attention-layer}
\end{equation}
where \(\sigma\) is row-wise softmax and \(M\) prevents attention
to the query token as a key. The model is trained by squared prediction
loss, without any supervision on conditioning. Training details are
given in Appendix~\ref{app:training-details}.

\noindent\textbf{Layerwise symmetry intervention.}
The linear system has an orthogonal symmetry: for any \(U,V\in O(d)\), the group of \(d\times d\) orthogonal matrices, the transformation
\[
    \widetilde X=UXV,\qquad
    \widetilde y=Uy,\qquad
    \widetilde x_q=V^\top x_q
\]
preserves the problem, since
\(\widetilde X(V^\top w)=\widetilde y\) and
\(\widetilde x_q^\top(V^\top w)=x_q^\top w\).
Thus the same prediction problem can be represented in different example and feature coordinates.

To make the learned computation identifiable, we adapt the layerwise
symmetry intervention of \citet{lutz2026layerwise}. At each layer, we sample independent \(U_\ell,V_\ell\sim\operatorname{Haar}(O(d))\), extend them to fix the query and label coordinates as
\[
    \bar U_\ell=
    \begin{bmatrix}U_\ell&0\\0&1\end{bmatrix},
    \qquad
    \bar V_\ell=
    \begin{bmatrix}V_\ell&0\\0&1\end{bmatrix}.
\]
and evaluate attention in the transformed coordinates before mapping the residual update back:
\begin{equation}
    Z_{\ell+1}
    =
    Z_\ell+
    \bar U_\ell^\top
    \operatorname{Attn}_\ell
    \!\left(
        \bar U_\ell Z_\ell\bar V_\ell
    \right)
    \bar V_\ell^\top .
    \label{eq:intervened-attention-layer}
\end{equation}
This is the transformer architecture we train and analyze throughout this section. The intervention discourages reliance on arbitrary coordinate choices, making the learned computation easier to identify, without imposing any conditioning objective or preconditioning algorithm.

%% file: paper/emergence2.tex
\subsection{Extracting a conditioning primitive}
\label{sec:algorithmic_template}
\label{subsec:row-side-preconditioner}

The transformer achieves increasingly accurate predictions with depth, following an error profile distinct from GD and CG at matched iteration budgets (Appendix~\ref{app:behavioral-diagnostics}). This suggests that the model learns a distinct, nontrivial procedure rather than simply unrolling either classical solver. We therefore examine its trained layers to identify the underlying computation.

\begin{figure}[t]
    \centering
    \includegraphics[width=0.61\linewidth]{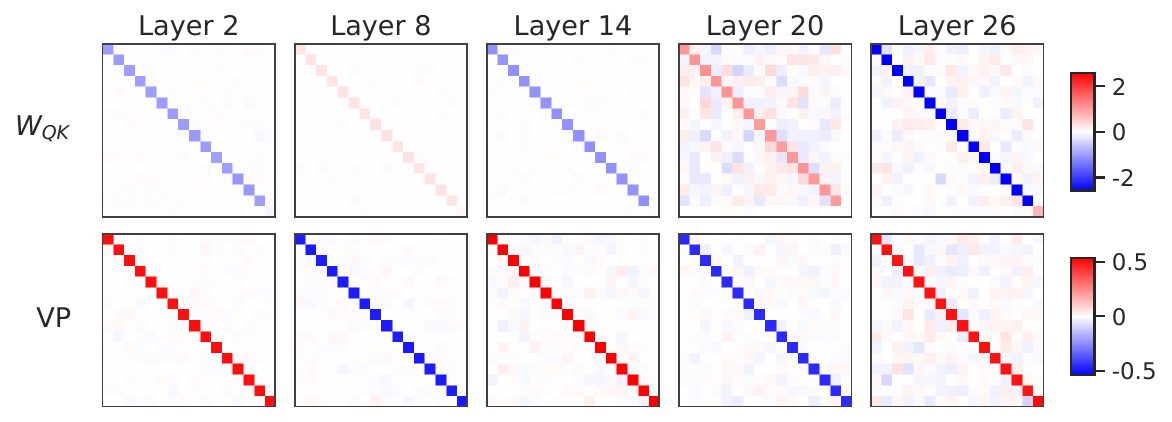}\hfil
    \includegraphics[width=0.33\linewidth]{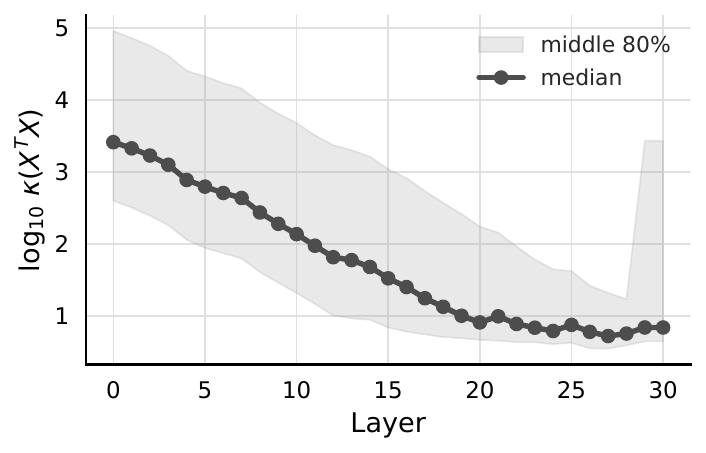}
    \caption{\textbf{Extracting a conditioning primitive.}
    \textit{Left:} Representative learned weight products
    \(W_{QK,\ell}\) and \(W_{VP,\ell}\), showing the dominant
    two-scalar structure in Eq.~\eqref{eq:weight_template}.
    \textit{Right:} Median condition number of the context Gram
    matrix across layers of the trained transformer, decreasing
    from roughly \(3{,}000\) to below \(10\) at the best layer.}
    \vspace{-\baselineskip}
    \label{fig:extraction}
\end{figure}

\noindent\textbf{Identifying the learned update.} The attention layers depend on the weights only through the products
\(W_{QK,\ell}=W_{Q,\ell}W_{K,\ell}^\top\) and
\(W_{VP,\ell}=W_{V,\ell}W_{P,\ell}\).
These products exhibit a simple dominant structure:
\(W_{QK,\ell}\) is concentrated on the feature-feature identity block,
while \(W_{VP,\ell}\) is approximately a scalar identity
(Figure~\ref{fig:extraction}, left).
We therefore approximate each layer by
\begin{equation}
    W_{QK,\ell}\approx
    \gamma_\ell
    \begin{bmatrix}
        I_d & 0\\
        0 & 0
    \end{bmatrix},
    \qquad
    W_{VP,\ell}\approx \alpha_\ell I_{d+1},
    \label{eq:weight_template}
\end{equation}
where \(\gamma_\ell\) and \(\alpha_\ell\) are the mean diagonal
entries of the respective identity blocks.

Substituting these approximated weight products into the intervened transformer
gives an explicit iteration. Writing \(X_\ell,y_\ell\) for the context
state and \(G_\ell=X_\ell X_\ell^\top\) for its Gram matrix, each layer
computes
\begin{equation}
    K_\ell
    =
    \sigma\!\left(
        \gamma_\ell U_\ell G_\ell U_\ell^\top
    \right),
    \qquad
    C_\ell=I+\alpha_\ell U_\ell^\top K_\ell U_\ell,
    \label{eq:projected_attention}
\end{equation}
and updates examples and labels by the same row transformation:
\begin{equation}
    X_{\ell+1}=C_\ell X_\ell,
    \qquad
    y_{\ell+1}=C_\ell y_\ell.
    \label{eq:abstract_context_iteration}
\end{equation}
To assess the approximation, we replace the learned weight products
in every layer by their two-scalar forms in Eq.~\eqref{eq:weight_template} and compare predictions
against the trained transformer on the same held-out tasks, using
identical random rotations. The predictions agree closely
(\(R^2>0.98\)), with a test MSE change of only
\(6.4\times10^{-3}\) (Appendix~\ref{app:extracted_iteration_derivation}). This confirms that the two-scalar update faithfully reproduces the trained transformer's predictive behavior.

\noindent\textbf{A solution-preserving congruence update.}
The key property of Eq.~\eqref{eq:abstract_context_iteration} is
that it preserves the linear-system relation:
if \(y_\ell=X_\ell w\), then
\(y_{\ell+1}=C_\ell X_\ell w=X_{\ell+1}w\).
When \(C_\ell\) is invertible, the transformed system has the
same solution. Moreover, the Gram matrix evolves by congruence:
\begin{equation}
    G_{\ell+1}
    =
    C_\ell G_\ell C_\ell^\top.
    \label{eq:extracted-conditioning-update}
\end{equation}
Crucially, \(C_\ell\) depends only on \(G_\ell\) and the random
rotation \(U_\ell\), not on the labels or query. Thus the extracted
iteration defines a self-contained conditioning mechanism on SPD
matrices, independent of the original prediction task.

\noindent\textbf{Empirical conditioning and the learned sign rule.}
The trained dynamics substantially improve conditioning: the median
condition number of \(G_\ell\) decreases from roughly \(3{,}000\)
to below \(10\) at the best layer (Figure~\ref{fig:extraction},
right). The two-scalar approximation exhibits similar behavior.

Which layers improve conditioning? We find that layers satisfying
\begin{equation}
    \operatorname{sign}(\alpha_\ell)
    \neq
    \operatorname{sign}(\gamma_\ell),
    \qquad
    |\alpha_\ell|<1
    \label{eq:conditioning-parameter-rule}
\end{equation}
typically reduce the condition number. Across runs, \(80\%\) of
layers satisfy this rule, and in the median such layer, \(95.5\%\)
of test instances experience a condition-number decrease
(Appendix~\ref{app:stepsize-conditions}).

Together, these results identify a Gram-dependent conditioning primitive within the transformer's learned prediction procedure. The symmetry intervention provides random coordinate mixing, while the learned attention determines the row corrections. The next section develops this primitive into RAPID by replacing dense rotations and soft attention with efficient structured mixing and sparse updates, yielding a standalone preconditioner without learned components at solve time.

%% file: paper/practical2.tex
\section{RAPID: Randomized Anytime Preconditioning by Iterative Dynamics}
\label{sec:practical}

Section~\ref{sec:setup-extracted-preconditioner} identified a
Gram-dependent congruence update learned by a transformer trained
only for prediction. We now develop this primitive into RAPID,
a standalone preconditioner for explicit SPD systems \(Ax=b\).
RAPID retains random coordinate mixing and row-wise corrections,
but replaces dense rotations and soft attention with efficient
structured transforms and sparse updates. It requires no trained
model or learned parameters at solve time.

\subsection{Constructing the RAPID update}

\noindent\textbf{Randomized mixing.}
Starting from \(A_0=A\), RAPID repeatedly applies an orthogonal
change of coordinates,
\[
    \widetilde A_t=Q_tA_tQ_t^\top.
\]
This preserves the eigenvalues, but exposes different off-diagonal
interactions that can be corrected locally. To obtain this mixing
efficiently, RAPID replaces dense random rotations with products
of randomized Walsh--Hadamard transforms:
\begin{equation}
    Q_t=HS_{t,k}\cdots HS_{t,2}HS_{t,1},
    \label{eq:rapid-mixer}
\end{equation}
where \(H\) is the normalized Walsh--Hadamard matrix and
\(S_{t,j}\) are independent diagonal matrices with random
\(\pm1\) entries. Such transforms spread information across
coordinates similarly to dense random rotations while
preserving orthogonality
\citep{ailon09fast,le2013fastfood,yu2016orthogonal}.
For fixed \(k\), they cost \(O(d\log d)\) per vector,
rather than \(O(d^2)\) for a dense rotation.

\noindent\textbf{From learned attention to sparse row corrections.}
The learned conditioning rule~\eqref{eq:conditioning-parameter-rule}
identifies opposite signs for \(\alpha_\ell\) and \(\gamma_\ell\),
with \(|\alpha_\ell|<1\). RAPID fixes the convention of positive
residual steps \(\alpha_\ell>0\), corresponding to negative attention scales \(\gamma_\ell<0\) in
the learned model; the opposite convention analogously motivates
negative steps with row-maximum selection.

With a negative attention scale, softmax favors smaller entries
of the mixed Gram matrix. For computational efficiency, RAPID
replaces dense soft attention with selection of the smallest
off-diagonal entry in each row:
\begin{equation}
    \tau_i(B):=\arg\min_{j\ne i}B_{ij},
    \qquad
    \mathcal R(B)_{ij}
    :=\mathbf 1\{j=\tau_i(B)\}.
    \label{eq:rapid-row-min}
\end{equation}
We exclude the diagonal to target interactions between different
rows rather than simply rescaling individual rows. The resulting
one-sparse-per-row operator reduces the cost of the congruence
update from \(O(d^3)\) to \(O(d^2)\). This sparsification is a
design choice motivated by the learned attention.

Writing \(R_t=\mathcal R(\widetilde A_t)\), RAPID applies
\begin{equation}
    C_t=I+\alpha_tR_t,
    \qquad
    A_{t+1}=C_t\widetilde A_tC_t^\top.
    \label{eq:rapid-update}
\end{equation}
This retains the congruence structure of the learned iteration,
now with an explicit sparse row operator.

\noindent\textbf{Why does this improve conditioning?}
The intuition is clearest for a matrix with a single exceptional
eigenvalue. Consider
\(\widetilde A=I+(\lambda-1)vv^\top\), where
\(\lambda>0\), \(\lambda\neq1\), and \(\|v\|=1\), so all
eigenvalues equal \(1\) except \(\lambda\).
Adding a small positive fraction \(\alpha\) of row \(j\) to
row \(i\), together with the corresponding column operation,
changes the exceptional eigenvalue to first order by
\begin{equation}
    \lambda(\alpha)
    =\lambda+\frac{2\alpha\lambda}{\lambda-1}
    \widetilde A_{ij}+O(\alpha^2).
    \label{eq:intuition}
\end{equation}
Thus, for sufficiently small \(\alpha>0\), a negative
\(\widetilde A_{ij}\) moves \(\lambda\) toward \(1\):
decreasing it when \(\lambda>1\) and increasing it when
\(\lambda<1\). Random mixing can expose negative off-diagonal entries even when none are present initially, and row-minimum selection favors such entries when available.
For general matrices, individual corrections need not improve
conditioning; Section~\ref{sec:theory} establishes a global
guarantee for an idealized variant. The first-order calculation
is detailed in Appendix~\ref{app:eigenvalue-correction}.

\noindent\textbf{Step-size selection.}
Each row of \(R_t\) selects one column, but multiple rows
may select the same column. Let
\[
    m_t=\max_j\sum_i(R_t)_{ij}
\]
denote the maximum column load. Since
\(\|R_t\|_{\mathrm{op}}=\sqrt{m_t}\), we use
\begin{equation}
    \alpha_t=\frac{s_\alpha}{\sqrt{m_t d\log d}},
    \label{eq:rapid-stepsize}
\end{equation}
where \(s_\alpha>0\) is a tunable scale. This keeps the
overall update magnitude
\(\alpha_t\|R_t\|_{\mathrm{op}}
=s_\alpha/\sqrt{d\log d}\)
independent of the column load, controlling the effect of simultaneous row corrections. Section~\ref{sec:theory} provides a complementary motivation based on the expected improvement in conditioning.

\subsection{RAPID as an anytime preconditioner}

\noindent\textbf{Accumulating the preconditioner.}
RAPID maintains a factor \(P_t\) satisfying
\begin{equation}
    P_0=I_d,\qquad
    P_{t+1}=C_tQ_tP_t,\qquad
    A_t=P_tAP_t^\top.
    \label{eq:rapid-factor}
\end{equation}
Thus the random basis changes are accumulated rather than
reversed after each correction.

To ensure a valid preconditioner at every iteration, we choose
\(s_\alpha\) so that \(\alpha_t<1\). Indeed, \(R_t\) is
row-stochastic, so its eigenvalues have magnitude at most \(1\).
Hence \(C_t=I+\alpha_tR_t\) is invertible whenever
\(0<\alpha_t<1\). Each congruence update therefore preserves
positive definiteness.

After any \(T\) iterations, RAPID produces the equivalent SPD
system
\[
    A_Tz=P_Tb,\qquad x=P_T^\top z,
\]
which can be solved using conjugate gradients. Since every
iteration yields a valid factor \(P_t\), RAPID is an anytime
preconditioner: the iteration budget trades setup cost against
the conditioning achieved. The factor can also be reused for
multiple right-hand sides.

\noindent\textbf{Complete algorithm.}
Algorithm~\ref{alg:rapid} summarizes the practical RAPID
iteration, using structured orthogonal mixing and row-minimum selection.

\begin{algorithm}[t]
\caption{RAPID: Practical preconditioning algorithm}
\label{alg:rapid}
\begin{algorithmic}[1]
\State \textbf{Input:} \(A\succ0\), iterations \(T\),
mixer depth \(k\), scale \(0<s_\alpha<\sqrt{d\log d}\)
\State \(A_0\gets A,\quad P_0\gets I_d\)
\For{\(t=0,\ldots,T-1\)}
    \State Draw \(Q_t\) via~\eqref{eq:rapid-mixer};
    \(\widetilde A_t\gets Q_tA_tQ_t^\top\)
    \Comment{random mixing}
    \State \(R_t\gets\mathcal R(\widetilde A_t)\) defined as in \eqref{eq:rapid-row-min}
    \Comment{row selection}
    \State \(m_t\gets\max_j\sum_i(R_t)_{ij}\); \(\alpha_t\gets s_\alpha/\sqrt{m_td\log d}\)\Comment{compute step-size}
    \State \(C_t\gets I+\alpha_tR_t\)
    \Comment{sparse correction}
    \State \(A_{t+1}\gets C_t\widetilde A_tC_t^\top\); \(P_{t+1}\gets C_tQ_tP_t\)
    \Comment{update operator}
\EndFor
\State \Return \(P_T, A_T\)\Comment{verifies $A_T=P_TAP_T^\top$}
\end{algorithmic}
\end{algorithm}

\noindent\textbf{Computational cost and theoretical scope.}
Row-minimum selection and the sparse congruence update each
cost \(O(d^2)\). Applying a fixed number of Walsh--Hadamard
transforms to a dense matrix costs \(O(d^2\log d)\), which
dominates the per-iteration complexity. These operations are
highly parallelizable and require no matrix factorization.
For dimensions that are not powers of two, the implementation
uses padding (Appendix~\ref{appx:simulation-details}).

Section~\ref{sec:experiments} evaluates the practical algorithm, while Section~\ref{sec:theory} analyzes a related variant using Haar mixing and block-restricted row selection.

%% file: paper/simulations2.tex
\section{Experimental Evaluation}
\label{sec:experiments}

In this section, we show that RAPID improves conditioning across SPD matrices
well beyond the transformer's training distribution, and that
these improvements translate into faster CG convergence. We use
the practical algorithm of Section~\ref{sec:practical}, with
Walsh--Hadamard mixing and global row-minimum selection. No
trained model is used in these experiments.

\noindent\textbf{Experimental protocol.}
We generate synthetic matrices $A=U\Lambda U^\top$, with Haar-random
$U$, dimension $d=1024$, and condition number $\kappa=10^4$.
We consider four distinct spectral distributions: Chebyshev,
linear, logarithmic, and power-law, with $100$ independent matrices
per family. We also evaluate $31$ SPD matrices from SuiteSparse
\citep{kolodziej2019suitesparse}, selected independently of RAPID's
performance by dimension ($100\le d\le6000$) and post-Jacobi
condition number ($10^3$--$10^6$). All matrices are treated as
dense operators. Details appear in
Appendix~\ref{appx:simulation-details}.

After Jacobi scaling, we compare RAPID ($s_\alpha=10$) with
randomized Nystr\"om preconditioning
\citep{frangella2023randomized}, which uses a one-shot low-rank
approximation rather than RAPID's iterative congruence updates
(sketch sizes $1\%$--$50\%$ of $d$).
We report condition numbers relative to the Jacobi-scaled
matrix $A_0$ and CG iterations relative to Jacobi-preconditioned
CG. All methods use the same right-hand side and stopping criterion
$\|A_0\widehat x-b_0\|_2/\|b_0\|_2\le0.002$, evaluated in
the original Jacobi-scaled coordinates, ensuring a common
accuracy target.

\begin{figure*}[t]
    \centering
    \includegraphics[width=\textwidth]{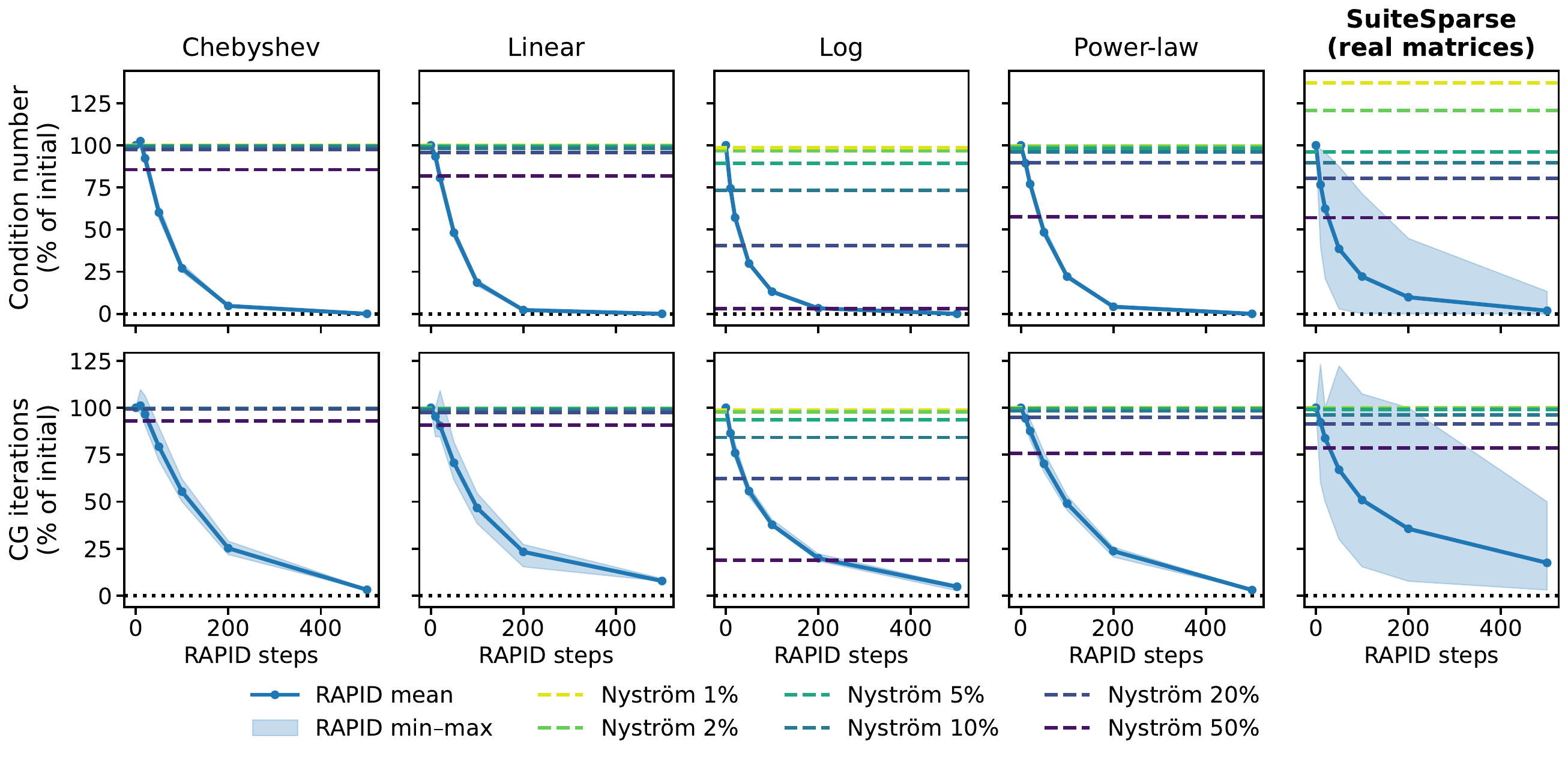}
    \caption{\textbf{Conditioning and CG convergence across synthetic
    and real-world SPD matrices.}
    Top: condition number relative to the Jacobi-scaled system.
    Bottom: CG iterations required to reach the common residual
    tolerance, relative to Jacobi-preconditioned CG.
    Lower is better in both rows. Blue curves show RAPID means
    and shaded bands the minimum and maximum; dashed lines show
    mean Nystr\"om results at different sketch sizes.
    Each synthetic panel contains $100$ independent matrices;
    SuiteSparse contains $31$. 
    The corresponding absolute values are reported in
    Appendix~\ref{appx:simulation-details}.
    }
    \label{fig:rapid-condition-relative}
\end{figure*}

\noindent\textbf{Broad improvements in conditioning.}
Figure~\ref{fig:rapid-condition-relative} (top) shows that RAPID
substantially improves conditioning across all four spectral
families. In contrast, Nystr\"om is strongly dependent on the
eigenvalue distribution: it is particularly effective on the
logarithmic spectrum, but achieves much smaller improvements on
Chebyshev and linear spectra, even with large sketches.
Remarkably, at every tested step count $T$, RAPID reduces
the condition number of \emph{all 31} SuiteSparse matrices,
not just the average.
These results demonstrate that RAPID generalizes well beyond
the Gaussian Gram matrices used during transformer training.

\noindent\textbf{Fewer CG iterations at fixed accuracy.}
Figure~\ref{fig:rapid-condition-relative} (bottom) shows that
RAPID's conditioning gains also translate into substantially
fewer CG iterations to reach the same accuracy, across all four
synthetic families and, on average, across SuiteSparse matrices.
This is significant because CG convergence depends on the full
spectrum, not just the condition number. Thus, RAPID not only
improves conditioning broadly, but also makes these systems
easier to solve iteratively. All solvers reached the target accuracy, except for Nystr\"om preconditioning with a $1\%$ sketch which failed on a single SuiteSparse instance.

\noindent\textbf{Practical considerations.}
The progressive improvements with increasing $T$ illustrate
RAPID's anytime nature, allowing a trade-off between setup
cost and preconditioner quality. A RAPID step costs
$O(d^2\log d)$, matching a dense CG iteration up to logarithmic
factors. In practice, RAPID can be more expensive per step,
as its operations do not benefit from the same highly optimized
GPU implementations available for CG's matrix--vector products.
The setup cost may nevertheless pay off through fewer CG
iterations, especially when amortized across multiple right-hand
sides. A definitive runtime comparison requires optimized
implementations of both methods and is left for future work.

%% file: paper/analysis3.tex
\section{Analyzing the Extracted Preconditioner} \label{sec:theory}

We analyze an idealized version of RAPID using two
simplifications relative to Algorithm~\ref{alg:rapid}:
(i) Haar-random mixing, which enables an exact calculation
of a certain expected `row-min drift' term, and (ii) block-restricted
row selection, which bounds the maximum column load of $R_t$.
These assumptions enable our proof, but are not required
by the practical algorithm. We first establish the
conditioning guarantee, then examine the empirical effect
of both simplifications.

\subsection{Conditioning guarantee under simplifying assumptions}

Our proof tracks a smooth approximation to the log-condition
number and shows that it decreases in expectation until the
matrix becomes well-conditioned. The key is to balance two
effects: the favorable first-order drift induced by row-min
selection, and a second-order error that depends on how many
rows select the same column. Haar mixing allows us to
characterize the drift, while blocking controls the error.

Specifically, we take $Q_t\sim\mathrm{Haar}(O(d))$ and
replace the global row-min operator $\mathcal R$ with a
block-restricted version. Partition the indices into blocks
of size at most $b$, and let $B(i)$ denote the block containing
$i$. Define
\begin{equation}\label{eq:block_min}
\tau_i^b(A):=\arg\min_{j\in B(i)\setminus\{i\}} A_{ij},
\qquad
\mathcal R_b(A)_{ij}:=\mathbf 1\{j=\tau_i^b(A)\}.
\end{equation}
Here $3\le b\le d$ is the block size; setting $b=d$
recovers the original global row-min operator $\mathcal R$.
By restricting selection to each row's block,
we ensure that no column is selected more than $b-1$
times. We now establish the two bounds and combine them
to obtain the conditioning guarantee. Full proofs appear
in Appendix~\ref{app:haar-block-row-min-proof}.

\newcommand{\tA}{\widetilde{A}} \newcommand{\tO}{\widetilde{O}}

\textbf{Potential Function \& First-order analysis.} To measure progress toward a well-conditioned matrix, we construct the following potential function:
\[\Phi_\beta(A) := \frac{1}{\beta} \log \frac{\tr(A^\beta)}{d} + \frac{1}{\beta} \log \frac{\tr(A^{-\beta})}{d}, \] where $\beta \ge 1$. This scale and basis invariant potential essentially corresponds to the log of the ratio of a softmax and softmin of the spectrum (with temperature $\beta$) and satisfies \[ 0 \le \Phi_\beta(A) \le \log \kappa(A) \le \Phi_\beta(A) + \frac{2\log d}{\beta}. \] We will set $\beta = \Theta(\log(d))$ below, making the approximation error constant. The key lemma capturing the decay of this potential arises from a Taylor expansion, where the curvature term is controlled by submultiplicativity of Schatten $p$-norms.
\begin{mylemma}\label{lemma:maintext_taylor}
    Let $m_t := \|R_t\|_1 = \max_{j} \sum_i (R_t)_{ij}$ be the maximal column load of the matrix $R_t$. If $\alpha \sqrt{m_t} \le 1/2,$ then \[ \Phi(A_{t+1}) \le \Phi(A_t) + 2\alpha f(R_t, \tA_t)  + 12\alpha^2 \beta m_t\]
    where \[f(R_t, \tA_t) :=   \left( \frac{\tr(R_t \tA_t^\beta)}{\tr(\tA_t^\beta)} - \frac{\tr(R_t \tA_t^{-\beta})}{\tr(\tA_t^{-\beta})} \right).\]
\end{mylemma}

\textbf{Row-min drift.} The key question is whether the first-order term $f(R_t,\tA_t)$ is negative in expectation, so that it outweighs the second-order growth. Haar mixing allows us to characterize this term exactly.
\begin{mylemma}\label{lemma:maintext_drift}
    Suppose $Q_t \sim \mathrm{Haar}(O(d))$ independently of the history, and that $R_t = \mathcal{R}_b(\tA_t)$. Then \begin{equation} \label{eq:maintext_drift} \mathbb{E}\left[ f(R_t, \tA_t) \middle| A_t \right] = \frac{-\nu_{b,d}}{\sqrt{d-1}} S_\beta(A_t), \end{equation} where $\nu_{b,d} \ge \sqrt{\log(b)},$ and $S_\beta$ depends on the spectral properties of $A_t$ and $d$. In particular, if $\beta \ge \log(8d)/\log(8/7)$ and $\kappa(A_t) \ge 2,$ then $S_\beta(A_t) \ge 1/4.$ 
\end{mylemma}
$S_\beta(A_t)$ is a measure of the correlation of spectrum of $A$ with its $\beta$th power. The lemma guarantees a negative first-order drift of magnitude $\alpha \cdot \Omega(\smash{\sqrt{\log(b)/d}})$ when $\kappa(A_t)\gg1$. This bound is conservative: $S_\beta(A)$ can be as large as $\smash{\sqrt d}$ when $A$ or $A^{-1}$ has a highly spiked spectrum (Appendix~\ref{app:spiky}).

\textbf{Main Result.} With control on both the drift and the second order growth, as long as $\kappa(A_t) \ge 2$, and the column-load $m_t$ is not too large, we can observe a decay in the potential $\Phi_\beta$. Notice that this column load is directly controlled by using $\mathcal{R}_b,$ since due to the blocking structure, each column can have at most $b-1$ ones. Exploiting this yields a concrete bound on $\mathbb{E}[\Phi_\beta(A_t) - \Phi_{\beta}(A_0)],$ which allows us to control the expected time to conditioning.
\begin{theorem}\label{thm:maintext_conditioning_haar}
    Suppose $Q_t \sim \Haar(O(d))$, and $R_t = \mathcal{R}_b(\tA_t),$ with $b = O(\log(d))$. Let $\beta_* := \log(d)/\log(8/7),$ and set $\alpha = \nu_{b,d}/\beta_*(b-1) \sqrt{d-1}.$ Let $\tau_* := \inf\{t : \kappa(A_t) \le 2\}$. Then \[ \mathbb{E}[\tau_*] = O\left( d \cdot \mathrm{polylog}(d) \cdot \log(\kappa(A_0))\right).\] 
\end{theorem}
Thus, the idealized RAPID variant achieves conditioning in $\tO(d)$ expected iterations, with logarithmic dependence on the initial condition number (see Appendix~\ref{app:haar-block-row-min-proof}). The proof uses the pessimistic bound $S_\beta\ge1/4$; more favorable spectra can yield faster conditioning and permit larger step sizes.

\textbf{Practical step size without blocking.} Balancing the negative drift of Lemma~\ref{lemma:maintext_drift} against the second-order term of Lemma~\ref{lemma:maintext_taylor} suggests that we should choose $\alpha$ to  $ \alpha_* = \smash{\frac{\nu_{b,d}}{12 \beta \cdot \sqrt{d-1} } \cdot \frac{S_\beta}{m_t}}$ to maximize the expected gain. While $m_t$ is easily computed, $S_\beta$ depends on the spectrum and is difficult to estimate. The proof therefore controls $m_t$ through blocking and uses a worst-case bound on $S_\beta$. Empirically, however, we observe $S_\beta/\sqrt{m_t}=\Omega(1)$ even with global row selection, across diverse spectra and dimensions (Appendix~\ref{app:drift-m-bound}). This suggests the practical step-size rule $\alpha_*'=\Theta(1/\sqrt{\beta m_t d})$, which we use in our experiments.

\noindent\textbf{Beyond Haar mixing.}
The Taylor bound of Lemma~\ref{lemma:maintext_taylor} and the spectral bound on $S_\beta$ do not depend on the mixing distribution. Haar mixing is used only to establish the exact expected-drift identity \eqref{eq:maintext_drift}, exploiting its rotational symmetry to characterize the effect of row-min selection. Randomized Walsh--Hadamard transforms also spread information across coordinates, but lack full rotational symmetry, making the drift harder to characterize. Extending the guarantee to randomized Walsh--Hadamard mixing would therefore require a corresponding drift bound, which remains open (see Appendix~\ref{subsec:fast-hadamard-condition}).

\subsection{Empirical role of the simplifying assumptions}

Theorem~\ref{thm:maintext_conditioning_haar} assumes Haar mixing
to compute the expected drift and block-restricted row minima
to bound the column load. We test whether these assumptions
also improve empirical conditioning. Both comparisons use
synthetic matrices with $d=2048$ and initial $\kappa=10^4$.

\noindent\textbf{Fast mixing.}
We compare Haar rotations with randomized Walsh--Hadamard
transforms, using identical initial matrices, global row
selection, and step-size rules. Across $100$ trials per spectral
family, Table~\ref{tab:rapid-mixing} shows that the final
condition numbers after $T=100$ steps differ by at most $0.5\%$.
Thus, fast structured mixing achieves essentially the same
conditioning as Haar mixing.

\noindent\textbf{Global versus block row selection.}
We compare the practical global selector \eqref{eq:rapid-row-min} with block-restricted
selection \eqref{eq:block_min} (block size $8$), using identical Walsh--Hadamard
mixing sequences. The global variant uses
$\alpha_t=s_\alpha/\sqrt{m_t d\log d}$, while the blocked
variant uses $\alpha_t=s_\alpha/\sqrt{d\log d}$.
Table~\ref{tab:rapid-block} shows that both improve
conditioning, but the practical configuration achieves
substantially larger reductions. Results are averaged over
$20$ trials; standard deviations are below $1.1$ percentage
points everywhere.

\begin{table*}[t]
\centering
\begin{minipage}[t]{0.48\textwidth}
\centering
\caption{\textbf{Haar versus Walsh--Hadamard mixing.}
Final condition numbers after $T=100$ steps (mean $\pm$ standard deviation).}
\label{tab:rapid-mixing}
\small
\begin{tabular}{lrr}
\toprule
Spectrum & Haar & Walsh--Hadamard\\
\midrule
Chebyshev & $5726\pm83$ & $5699\pm90$\\
Linear & $4397\pm77$ & $4397\pm82$\\
Log & $2848\pm27$ & $2844\pm25$\\
Power-law & $4555\pm58$ & $4564\pm62$\\
\bottomrule
\end{tabular}
\end{minipage}
\hfill
\begin{minipage}[t]{0.48\textwidth}
\centering
\caption{\textbf{Global versus block row minima.}
Mean condition-number reduction (\%).}
\label{tab:rapid-block}
\small
\setlength{\tabcolsep}{4pt}
\begin{tabular}{lrrrr}
\toprule
& \multicolumn{2}{c}{$T=100$}
& \multicolumn{2}{c}{$T=500$}\\
\cmidrule(lr){2-3}\cmidrule(l){4-5}
Spectrum & Global & Block & Global & Block\\
\midrule
Chebyshev & 43.1 & 1.0 & 98.8 & 26.6\\
Linear & 56.2 & 6.5 & 99.6 & 37.8\\
Log & 71.5 & 16.2 & 98.7 & 55.0\\
Power-law & 54.5 & 8.2 & 98.8 & 38.2\\
\bottomrule
\end{tabular}
\end{minipage}
\end{table*}

Together, these results show that the simplifying assumptions
used in our proof are not needed for effective conditioning
in practice: fast mixing matches Haar, while the practical
global configuration substantially outperforms blocking.
Establishing a theoretical guarantee for the practical
algorithm remains open.

%% file: paper/conclusion3.tex
\section{Conclusion}

We introduced RAPID, an anytime SPD preconditioner derived from a
conditioning mechanism discovered in a transformer trained for
in-context learning. The extracted update preserves the
linear-system relation and evolves its Gram matrix by congruence.
RAPID turns this mechanism into a standalone algorithm combining randomized mixing and sparse row corrections, accumulating a preconditioner $A_t=P_tAP_t^\top$. The transformer provides only the algorithmic primitive; RAPID requires neither model inference nor learned parameters at solve time.
For an idealized Haar-mixing, block-restricted variant, we prove
convergence to any fixed condition-number target for arbitrary SPD
matrices, with near-linear dependence on dimension and logarithmic
dependence on initial conditioning. Experiments with practical RAPID
show broad conditioning improvements and reduced CG iteration counts
on synthetic and SuiteSparse matrices.

Several challenges remain. Extending the convergence guarantee to
practical Walsh--Hadamard mixing and global row-minimum selection
would close the gap between theory and implementation. Establishing
competitive end-to-end runtime requires optimized implementations
and careful accounting of setup costs, particularly when amortized
across multiple right-hand sides. Finally, extending RAPID to sparse
and matrix-free systems will require avoiding explicitly formed
dense mixed matrices. More broadly, RAPID illustrates how studying
learned transformer computations can lead to independent numerical
algorithms that apply well beyond the original learning task.

\section*{Acknowledgments}
This research was supported by the National Science Foundation grant CPS-2317079.

%% file: paper/expts_appx.tex
\section{Supplementary material for Section~\ref{sec:setup-extracted-preconditioner}}

\subsection{Transformer architecture and training details}\label{app:training-details}

This appendix specifies the training setup used for the in-context regression
experiments in Section~\ref{subsec:algorithm-discovery-task}.  We train
attention-only transformers on noiseless synthetic linear regression instances.
Unless stated otherwise, the experiments use \(d=15\) covariates and
\(d+1=16\) tokens: \(d\) context examples and one query example.

\paragraph{Data distribution.}
For each batch element, we sample
\[
    x_1,\ldots,x_d,x_q \overset{\mathrm{i.i.d.}}{\sim} \mathcal N(0,I_d),
    \qquad
    w\sim \mathcal N(0,I_d),
\]
and set
\[
    y_i = x_i^\top w/\sqrt d,
    \qquad
    y_q = x_q^\top w/\sqrt d .
\]
The factor \(1/\sqrt d\) is a normalization convention; equivalently, the
effective regression coefficient has covariance \(I_d/d\).  The model input is
\[
    Z_0 =
    \begin{bmatrix}
        x_1^\top & y_1\\
        \vdots & \vdots\\
        x_d^\top & y_d\\
        x_q^\top & 0
    \end{bmatrix}
    \in \mathbb R^{(d+1)\times(d+1)} ,
\]
and the loss is the squared error on the query label only,
\[
    \mathcal L
    =
    \bigl((Z_L)_{d+1,d+1}+y_q\bigr)^2 .
\]
The implementation reads the prediction $y_q$ from the bottom-right entry of the final
token matrix with the same negative-output convention as
\citet{vonoswald23transformers}: \(\widehat y_q=-(Z_L)_{d+1,d+1}\).
Fresh inputs and fresh regression coefficients are resampled at every training
iteration.

\paragraph{Architecture.}
We use a residual attention-only transformer with no feed-forward blocks.  Each
layer has one attention head and embedding dimension \(d+1=16\).  For
\(Z\in\mathbb R^{(d+1)\times(d+1)}\), layer \(\ell\) computes
\[
    Z_{\ell+1}
    =
    Z_\ell
    +
    A_\ell(Z_\ell) Z_\ell W_{V,\ell}W_{P,\ell},
\]
where
\[
    A_\ell(Z)
    =
    \operatorname{softmax}
    \left(
        Z W_{Q,\ell} W_{K,\ell}^{\top} Z^\top
        + M
    \right).
\]
All linear projections are bias-free.  The matrices
\(W_{Q,\ell},W_{K,\ell},W_{V,\ell}\) are initialized with Xavier uniform
initialization, while the output projection \(W_{P,\ell}\) is initialized at
zero.  Thus the initial network is the identity map through the residual
connection. For notational simplicity, we absorb the \(1/\sqrt{d+1}\) factor commonly present in the attention architecture \citep{vaswani2017attention} into the attention weights.

The attention mask removes the query token as a key/value:
\[
    M
    =
    \begin{bmatrix}
        \mathbf 0_{(d+1)\times d} & -\infty\,\mathbf 1_{d+1}
    \end{bmatrix}.
\]
Thus every token may attend to the \(d\) labeled context examples, but no token
may attend to the query token.  The query row is therefore used only as the
prediction site.

\paragraph{Attention-logit clipping.}
For numerical stability and to avoid nearly one-hot attention early in
training, we clip attention logits after centering each attention row.  Let
\[
    s_{ij}
    =
    \frac{q_i^\top k_j}{\sqrt{d+1}}
\]
denote the unmasked attention score, and let \(\mathcal A_i\) be the set of allowed (non-masked) keys under the attention mask.  Before the softmax, we replace each
allowed score by
\[
    \widetilde s_{ij}
    =
    \max\left\{
        s_{ij}-\max_{j'\in\mathcal A_i}s_{ij'},
        -6
    \right\},
    \qquad j\in\mathcal A_i,
\]
and keep masked scores equal to \(-\infty\).  Hence the largest allowed logit in
each row is zero, and all allowed logits are lower bounded by \(-6\).  This
bounds the maximum probability ratio between two allowed tokens by \(e^6\). We found this clipping useful for numerical stability when training deep models.

\paragraph{Symmetry augmentation.}
The reported training runs use the symmetry-augmented implementation.  At each
layer and for each forward pass, we draw independent Haar orthogonal matrices
\(U\in\mathbb R^{d\times d}\) and
\(V\in\mathbb R^{d\times d}\), extend them by a fixed final
row/column, and apply the attention block in the transformed coordinates:
\[
    \widetilde Z_\ell
    =
    \operatorname{diag}(U,1)
    Z_\ell
    \operatorname{diag}(V,1).
\]
The attention update is then rotated back before being added residually to
\(Z_\ell\).  This augmentation preserves the distinguished query row and label
coordinate while randomizing the basis of the context examples and features.

\paragraph{Optimization.}
We train each model with AdamW for \(30{,}001\) steps using batch size \(8192\),
learning rate \(10^{-3}\), \((\beta_1,\beta_2)=(0.9,0.99)\), no weight decay,
and global gradient clipping at norm \(1\).  The learning rate is linearly
warmed up for the first \(500\) steps and then held fixed.  We resample both the
covariates and regression coefficient at every step.  We train ten independent
runs for each depth
\[
    L\in\{1,2,3,5,7,10,15,20,30\},
\]
and use the depth-\(30\) models for the extraction analysis unless otherwise
specified.

\begin{figure}[h]
    \centering
    \includegraphics[width=0.52\linewidth]{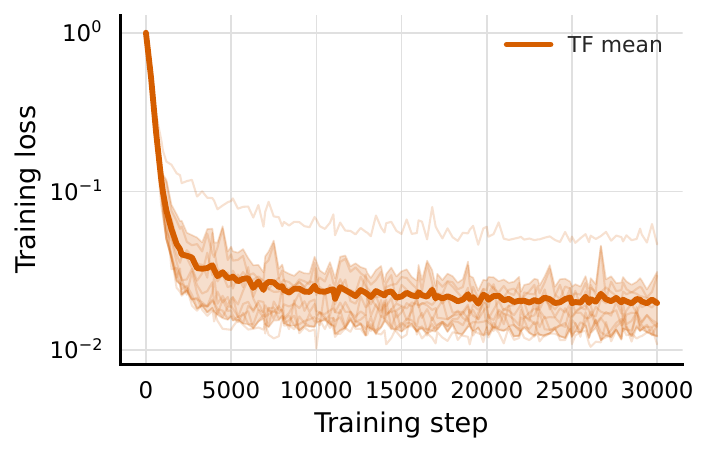}
    \caption{\textbf{Training curves for depth-\(30\) transformers.}
    The loss is the squared error on the query prediction.  Inputs and
    regression coefficients are resampled at every optimization step.}
    \label{fig:training-curves-depth30}
\end{figure}

\paragraph{Training curves.}
Figure~\ref{fig:training-curves-depth30} reports the training loss for the
depth-\(30\) models used in the extraction analysis.  Thin curves show
individual runs; the thick curve shows the mean across runs, with the shaded band denoting the central \(80\%\) interval.

\subsection{Behavioral diagnostics}
\label{app:behavioral-diagnostics}

We evaluate the trained symmetry-intervened transformers in the square \(d=15\) linear-system setting. For each depth \(L\le 30\), GD and CG are run for \(L\) iterations from zero. GD uses a fixed stepsize selected separately for each iteration budget from 60 logarithmically spaced values in \([10^{-4},3]\), minimizing query-prediction MSE on 4,096 validation tasks. Results are evaluated on 4,096 independent test tasks.

\begin{figure}[t]
    \centering
    \includegraphics[width=0.3\linewidth]{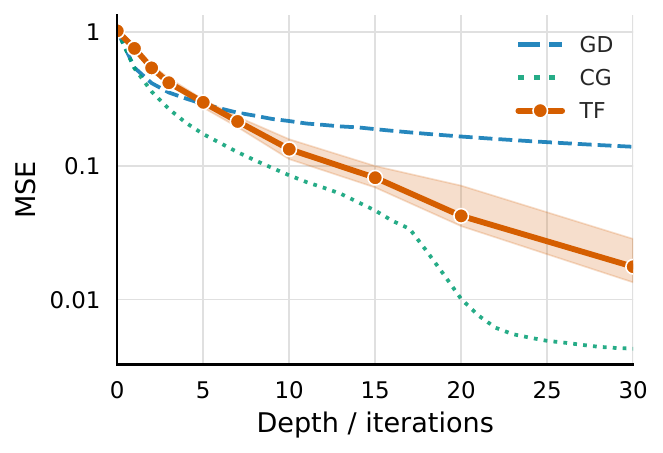}\hfil
    \includegraphics[width=0.6\linewidth]{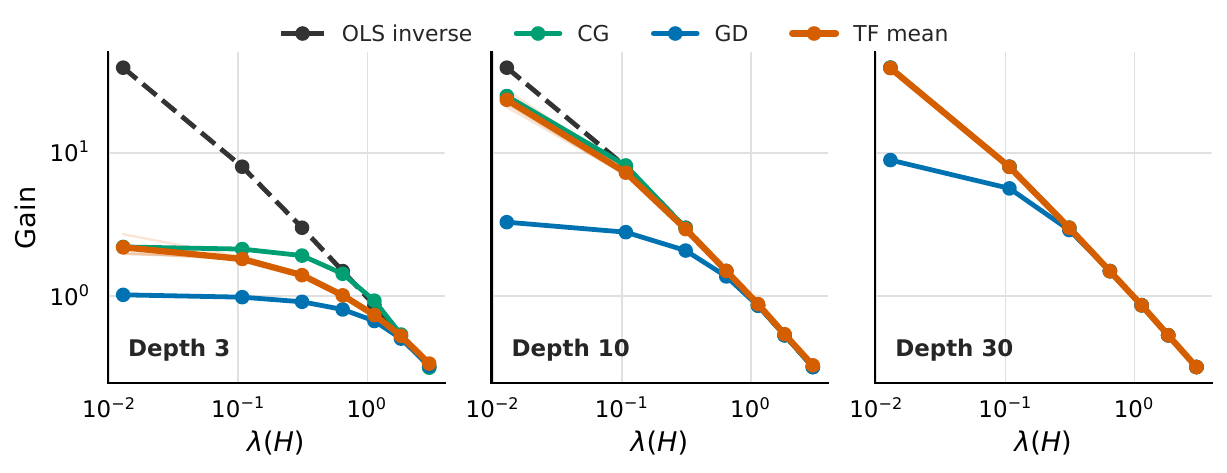}
    \caption{\textbf{Prediction behavior across transformer depths.}
    \textit{Left:} Test MSE decreases with model depth, following a
    profile distinct from GD and CG at matched iteration budgets.
    \textit{Right:} The fitted endpoint spectral gain approaches the
    OLS response \(1/\lambda\) across eigendirections of
    \(X^\top X/d\), including small eigenvalues.}
    \label{fig:transformer_solution}
\end{figure}

\paragraph{Prediction error.}
Figure~\ref{fig:transformer_solution} (left) shows that test MSE
decreases with model depth. The error profile differs from GD
and CG at matched iteration budgets, suggesting that the transformer
does not simply reproduce either classical solver.

\paragraph{Spectral inverse-kernel diagnostic.}
For each fixed context \(X\), we probe the endpoint with random queries and fit
the best linear query map
\[
    \widehat y_q^{\rm TF}
    \approx
    x_q^\top \widehat\beta_{\rm TF}(X).
\]
We then express \(\widehat\beta_{\rm TF}\) in the eigenbasis of
\(H=X^\top X/d=U\operatorname{diag}(\lambda_j)U^\top\). Since exact OLS
satisfies
\[
    \beta_{\star,e,j}=\lambda_j^{-1} b_{e,j},
    \qquad
    b_e=U^\top X^\top y/d,
\]
the ideal gain in eigendirection \(j\) is \(1/\lambda_j\). The plotted gain is
the zero-intercept slope mapping \(b_{e,j}\) to \(\widehat\beta_{e,j}\), after
pooling eigendirections across contexts and binning by \(\lambda_j\):
\[
    g_B
    =
    \frac{
        \sum_{(c,j)\in B}\widehat\beta_{e,c,j} b_{e,c,j}
    }{
        \sum_{(c,j)\in B} b_{e,c,j}^2
    } .
\]
Figure~\ref{fig:transformer_solution} (right) shows that the fitted
gain approaches the OLS response \(1/\lambda\) as depth increases,
including at small eigenvalues. Thus the transformer increasingly
captures the full inverse spectral response rather than only the
well-conditioned directions.

\begin{figure}[t]
    \centering
    \includegraphics[width=0.48\linewidth]{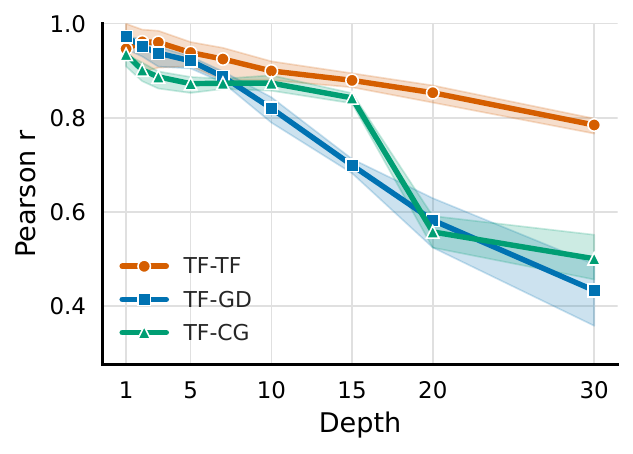}
    \hfill
    \includegraphics[width=0.48\linewidth]{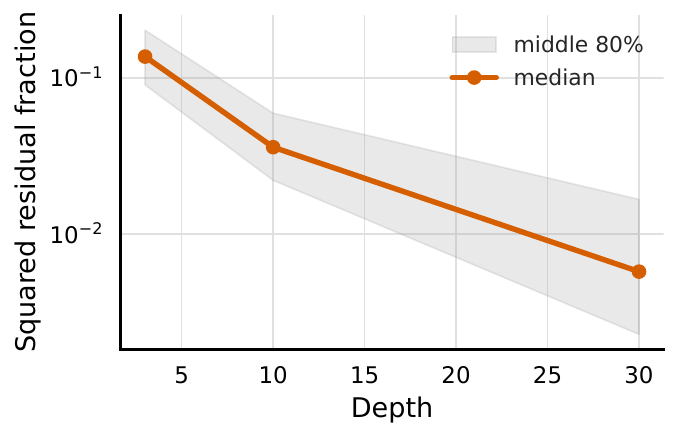}
    \caption{\textbf{Additional behavioral diagnostics.}
    \textit{Left:} Pearson correlations between centered residual-error
    vectors on shared test tasks. Transformer runs remain more aligned
    with each other than with GD or CG.
    \textit{Right:} Squared residual fraction of the best linear
    query-to-prediction map for each context. The small residual at
    depth \(30\) supports the spectral-filter interpretation.}
    \label{fig:behavioral-diagnostics-appx}
\end{figure}

\paragraph{Linearity of the query map.}
This spectral diagnostic is meaningful only if, for fixed context, the endpoint
is well approximated by a linear function of the query. We check this by
plotting the squared residual fraction
\[
    \frac{
        \sum_q
        \left(
            \widehat y_q^{\rm TF}
            -
            x_q^\top \widehat\beta_{\rm TF}
        \right)^2
    }{
        \sum_q
        \left(\widehat y_q^{\rm TF}\right)^2
    } .
\]
As shown in Figure~\ref{fig:behavioral-diagnostics-appx} (right), this residual
fraction decreases with depth and is below \(10^{-2}\) at depth \(30\). Thus
the fitted linear query map explains more than \(99\%\) of the transformer's
prediction energy on the probe queries, validating the spectral-filter
interpretation for the depth-\(30\) models.

\paragraph{Residual-error alignment.}
To compare failure modes, we compute the centered residual-error vector
\[
    e_c
    =
    (\widehat y-y_q)
    -
    \frac{1}{N}\mathbf 1\mathbf 1^\top(\widehat y-y_q)
\]
on a shared test set, and report Pearson correlations between such vectors.
Transformer-Transformer (TF-TF) alignment is computed by splitting trained transformer runs into two
groups and correlating the group-averaged residuals; TF-GD and TF-CG compare
the same transformer averages to GD and CG. As shown in
Figure~\ref{fig:behavioral-diagnostics-appx} (left), TF-TF alignment
remains high at depth \(30\) (\(r\approx 0.8\)), while TF-GD and
TF-CG fall below \(0.5\). Thus independently trained transformers share
a stable residual structure that is not well explained by either
classical baseline.

\subsection{Extracted iteration and fidelity diagnostics}
\label{app:extracted_iteration_derivation}

We describe the two-scalar approximation used in
Section~\ref{sec:algorithmic_template}.  For each trained layer, let
\[
    W_{QK,\ell}=W_{Q,\ell}W_{K,\ell}^{\top},
    \qquad
    W_{VP,\ell}=W_{V,\ell}W_{P,\ell}.
\]
We approximate these products as
\[
    \widehat W_{QK,\ell}
    =
    \gamma_\ell
    \begin{bmatrix}
        I_d & 0\\
        0 & 0
    \end{bmatrix},
    \qquad
    \widehat W_{VP,\ell}
    =
    \alpha_\ell I_{d+1}.
\]
Here \(\gamma_\ell\) is the mean diagonal entry of the feature-feature block of
\(W_{QK,\ell}\), and \(\alpha_\ell\) is the mean diagonal entry of the
feature-feature block of \(W_{VP,\ell}\).  In the projected value map, the label
coordinate is tied to the same scalar \(\alpha_\ell\), giving the two-scalar
template used in the main text.  All off-diagonal entries and all \(QK\) entries
involving the label coordinate are discarded.

Write the full state as
\[
    Z_\ell
    =
    \begin{bmatrix}
        X_\ell & y_\ell\\
        x_{q,\ell}^{\top} & \hat y_\ell
    \end{bmatrix}.
\]
Under the symmetry intervention, the layer applies attention in a randomly
rotated context basis and then rotates the update back.  With the projected
weights, the context attention is
\[
    A_\ell
    =
    \sigma
    \left(
        \gamma_\ell U_\ell X_\ell X_\ell^\top U_\ell^\top
    \right),
\]
and the corresponding context update is
\[
    X_{\ell+1}
    =
    C_\ell X_\ell,
    \qquad
    y_{\ell+1}
    =
    C_\ell y_\ell,
    \qquad
    C_\ell
    =
    I+\alpha_\ell U_\ell^\top A_\ell U_\ell .
\]
Thus the projection gives the row-side iteration used in the main text.  The
same projected layer also updates the query row and readout.  If
\[
    a_\ell^\top
    =
    \sigma
    \left(
        \gamma_\ell x_{q,\ell}^\top X_\ell^\top U_\ell^\top
    \right)
\]
is the query-to-context attention row, then
\[
    x_{q,\ell+1}^{\top}
    =
    x_{q,\ell}^{\top}
    +
    \alpha_\ell a_\ell^\top U_\ell X_\ell,
    \qquad
    \hat y_{\ell+1}
    =
    \hat y_\ell
    +
    \alpha_\ell a_\ell^\top U_\ell y_\ell .
\]

\begin{figure}[h]
    \centering
    \includegraphics[width=0.42\linewidth]{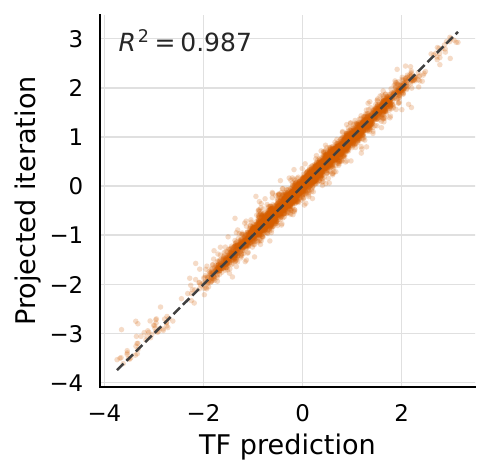}
    \caption{\textbf{Fidelity of the projected iteration.}
    Each point compares the original depth-\(30\) transformer prediction with
    the prediction from the two-scalar projected iteration on the same held-out
    task.  The high \(R^2\) shows that the low-dimensional projection captures
    the dominant predictive behavior of the trained model.}
    \label{fig:projected-prediction-scatter}
\end{figure}

\paragraph{Fidelity of the projection.}
We evaluate the projection by comparing the original depth-\(30\) transformer
and the projected iteration on the same held-out tasks, using the same random
rotations in both forward passes.  Predictions are pooled across evaluation
tasks and trained runs.  The reported \(R^2\) is the squared Pearson correlation
between the original transformer predictions and the projected-iteration
predictions:
\[
    R^2
    =
    \operatorname{Corr}
    \left(
        \widehat y^{\rm TF},
        \widehat y^{\rm proj}
    \right)^2 .
\]
The MSE change is computed on the same pooled predictions as
\[
    \Delta{\rm MSE}
    =
    \frac{1}{RN}
    \sum_{r=1}^{R}\sum_{i=1}^{N}
    \left[
        \left(\widehat y^{\rm proj}_{r,i}-y_i\right)^2
        -
        \left(\widehat y^{\rm TF}_{r,i}-y_i\right)^2
    \right].
\]
The projection preserves the trained model closely: \(R^2>0.98\), and the test
MSE changes by only \(6.4\times 10^{-3}\).  Figure~\ref{fig:projected-prediction-scatter}
shows the corresponding prediction-level agreement.

\subsection{Conditioning behavior and the layerwise sign rule}
\label{app:stepsize-conditions}

This appendix gives the empirical evidence behind the preconditioning
interpretation in
Section~\ref{subsec:row-side-preconditioner}.  The main questions are:
(i) does the extracted iteration actually improve conditioning layer by layer,
and (ii) when does a layer act as a conditioning step?

\begin{figure}[h]
    \centering
    \includegraphics[width=0.31\linewidth]{figures/depth30_condition.pdf}\hfill
    \includegraphics[width=0.31\linewidth]{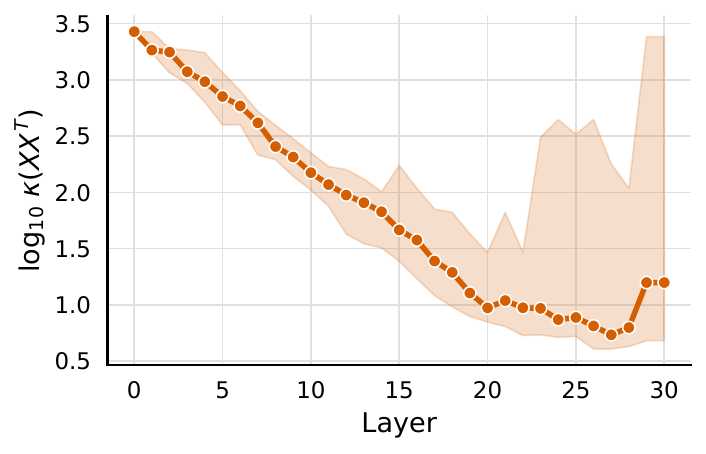}\hfill
    \includegraphics[width=0.31\linewidth]{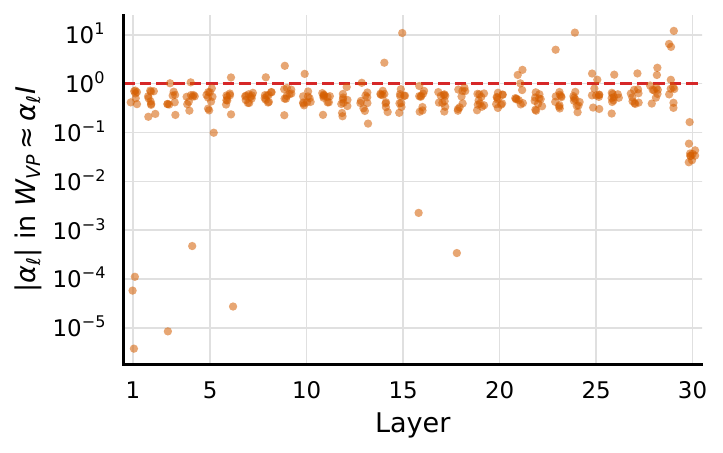}
    \caption{\textbf{Conditioning overview.}
    \textit{Left:} Condition number of \(G_\ell=X_\ell X_\ell^\top\)
    across layers of the trained transformer.
    \textit{Center:} The corresponding trajectory under the
    two-scalar approximation. Both reduce the median condition
    number from above \(3\times10^3\) to below \(10\) at the
    best layer.
    \textit{Right:} Distribution of learned step sizes
    \(\alpha_\ell\), most of which lie close to the sufficient
    invertibility threshold \(\alpha_\ell=1\).}
    \label{fig:conditioning-overview}
\end{figure}

\paragraph{Conditioning under the extracted and projected iterations.}
Figure~\ref{fig:conditioning-overview} compares the conditioning
trajectories of the trained depth-\(30\) transformer and its two-scalar
approximation. In both cases, the median condition number of
\(G_\ell=X_\ell X_\ell^\top\) decreases from above \(3\times10^3\)
to below \(10\) at the best layer. Thus the low-dimensional
approximation captures the conditioning behavior of the trained model.

The right panel shows that most learned step sizes \(\alpha_\ell\)
are close to \(1\). Since the softmax matrix \(K_\ell\) is
row-stochastic, \(|\alpha_\ell|<1\) guarantees that
\(I+\alpha_\ell K_\ell\) is invertible, preserving positive
definiteness under the congruence update. The learned values suggest
that the model prefers large residual steps within
this invertibility range.

Both trajectories exhibit occasional increases in condition
number, particularly near the final layers.  This is consistent with the transformer being trained for best final prediction at fixed depth, rather than for monotone conditioning at every intermediate layer.

\begin{figure}[h]
    \centering
    \includegraphics[width=0.8\textwidth]{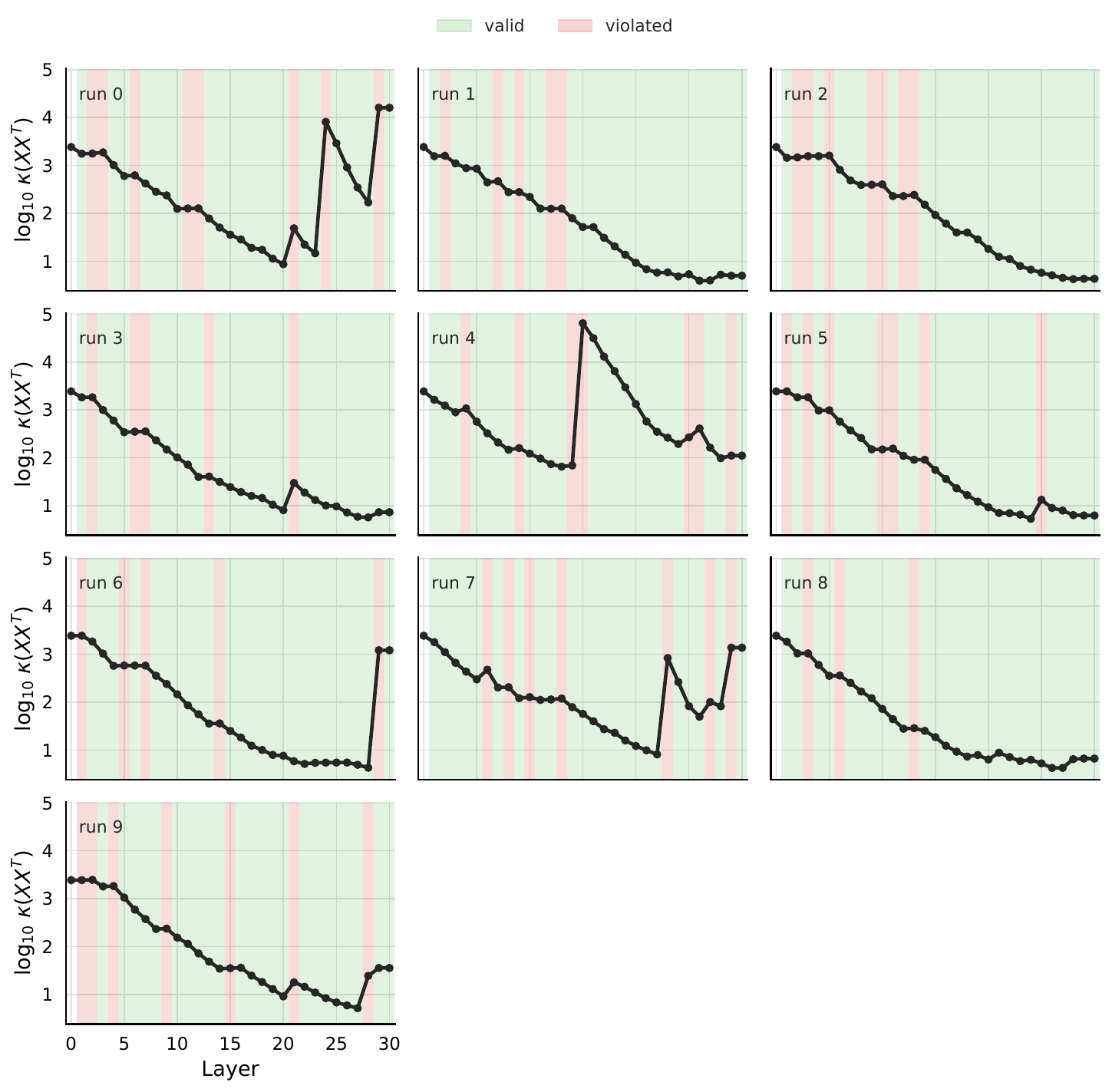}
    \caption{\textbf{Layerwise sign rule and conditioning.}
    Layerwise condition numbers compared with the empirical
    criterion
    \(\operatorname{sign}(\alpha_\ell)\neq
    \operatorname{sign}(\gamma_\ell)\) and
    \(|\alpha_\ell|<1\). Layers satisfying this rule typically
    decrease the condition number.}
    \label{fig:sign-rule-conditioning}
\end{figure}

\paragraph{Which layers condition?}
Empirically, layers tend to reduce the condition number when
their parameters satisfy
\begin{equation}
    \operatorname{sign}(\alpha_\ell)\neq\operatorname{sign}(\gamma_\ell),
    \qquad |\alpha_\ell|<1 .
    \label{eq:app-sign-rule}
\end{equation}
For example, when \(\alpha_\ell>0\) and \(\gamma_\ell<0\),
attention favors smaller entries of the mixed Gram matrix,
which receive positive weight in the residual update.
Figure~\ref{fig:sign-rule-conditioning} compares this rule
with the observed layerwise conditioning trajectories.

Across runs, \(80\%\) of layers satisfy the rule, and in the
median such layer, \(95.5\%\) of test instances decrease their
condition number. Thus the learned parameters provide a useful
empirical indicator of conditioning steps, although the rule
does not guarantee a decrease on every instance.

%% file: paper/simulations_appendix2.tex
\clearpage
\section{Additional Experimental Details}
\label{appx:simulation-details}

\noindent\textbf{Synthetic matrices.}
We generate $A=U\Lambda U^\top$, where $U$ is Haar-random
orthogonal and $\Lambda$ has one of four spectra, defined
for $i=0,\ldots,d-1$:
\[
\begin{aligned}
\text{Chebyshev:}\quad
&\lambda_i=\frac{\kappa+1}{2}
-\frac{\kappa-1}{2}\cos\frac{(2i+1)\pi}{2d},\\
\text{Linear:}\quad
&\lambda_i=1+(\kappa-1)\frac{i}{d-1},\\
\text{Logarithmic:}\quad
&\lambda_i=\kappa^{i/(d-1)},\\
\text{Power-law:}\quad
&\lambda_i=1+(\kappa-1)
\left(\frac{i}{d-1}\right)^2.
\end{aligned}
\]
We set the spectral endpoints to $1$ and $\kappa=10^4$
and use $d=1024$, with $100$ independent draws per family.

\noindent\textbf{SuiteSparse matrices.}
We use $31$ SPD matrices from the SuiteSparse Matrix
Collection \citep{kolodziej2019suitesparse}, selected by
dimension ($100\le d\le6000$) and post-Jacobi condition number
($10^3$--$10^6$), independently of RAPID's performance.
The selected instances, grouped by collection, are:

{\small
\begin{quote}
\texttt{Boeing}: msc00726, msc01440;

\texttt{Cannizzo}: sts4098;

\texttt{Cylshell}: s1rmq4m1, s1rmt3m1;

\texttt{HB}: 1138\_bus, 494\_bus, 662\_bus, 685\_bus,
bcsstk03, bcsstk04, bcsstk05, bcsstk06, bcsstk07,
bcsstk08, bcsstk09, bcsstk10, bcsstk13, bcsstk14,
bcsstk15, bcsstk21, bcsstk22, bcsstk26, bcsstk27,
bcsstm12, lund\_a, nos3, nos5;

\texttt{Nasa}: nasa1824, nasa2910;

\texttt{TKK}: plbuckle.
\end{quote}
}
All matrices are converted to dense form. We normalize by
an estimate of the largest eigenvalue (20 power iterations)
and apply Jacobi scaling. For each matrix, we sample an
independent Gaussian right-hand side before applying the
same coordinate scaling.

\noindent\textbf{RAPID implementation.}
We implement RAPID on the GPU using PyTorch/Triton and fast
Walsh--Hadamard transforms \citep{dao2024fast}.
We use one randomized Hadamard mixing stage ($k=1$),
global row-minimum selection, and $s_\alpha=10$.
A single mixing stage suffices in practice, as repeated
RAPID iterations progressively mix the coordinates.
The row-selection kernel finds the smallest off-diagonal
entry in each row and computes the maximum column
multiplicity $m_t$. A parallel Triton kernel then applies
the congruence update entrywise:
\[
\begin{aligned}
(A_{t+1})_{ij}
={}&\widetilde A_{ij}
+\alpha_t\widetilde A_{\tau(i),j}
+\alpha_t\widetilde A_{i,\tau(j)}
+\alpha_t^2\widetilde A_{\tau(i),\tau(j)},
\end{aligned}
\]
where $\alpha_t=s_\alpha/\sqrt{m_t d\log d}$.
We accumulate the transformation $P_t$ as in
Section~\ref{sec:practical} to recover solutions in the
original coordinates. Non-power-of-two dimensions are padded
to the next power of two with an identity block; the
right-hand side is padded with zeros. We evaluate RAPID
after $T\in\{10,20,50,100,200,500\}$ steps.

\noindent\textbf{Nystr\"om preconditioning.}
For each sketch fraction
$q\in\{0.01,0.02,0.05,0.1,0.2,0.5\}$, we use a Gaussian
sketch of rank $r=\lceil qd\rceil$, orthonormalized by QR.
We form $Y=A_0\Omega$ and $B=\Omega^\top Y$, then construct
a stabilized Nystr\"om approximation using Cholesky and
eigendecomposition. Its approximate eigenvectors $U$ and
positive eigenvalue estimates $\widehat\lambda_i$ define
the preconditioner
\[
M=I+U\operatorname{diag}
\left(\frac{\theta}{\widehat\lambda_i}-1\right)U^\top,
\qquad
\theta=\min_i\widehat\lambda_i.
\]
We apply $M$ within preconditioned CG.

\noindent\textbf{CG and evaluation.}
All solves use a common GPU float32 CG implementation,
initialized at zero, with at most $10^4$ iterations.
The true relative residual is checked every $10$ iterations,
after recovering the solution in the Jacobi-scaled reference
system $A_0x=b_0$. Convergence requires
\[
\frac{\|A_0\widehat x-b_0\|_2}{\|b_0\|_2}
\le 2\times10^{-3}.
\]
Condition numbers are computed from the extreme eigenvalues
of the transformed matrices. Figure~\ref{fig:rapid-condition-relative}
reports condition numbers and CG iteration counts relative
to the Jacobi baseline, with means and minimum-maximum
ranges for RAPID and means for Nystr\"om. All comparisons use
the same matrices and right-hand sides.

\begin{table*}[t]
\centering
\caption{\textbf{Absolute condition numbers and CG iterations.}
Mean values across $100$ independent matrices per synthetic
family and $31$ SuiteSparse matrices. All RAPID runs reached
the target accuracy ($100/100$ per synthetic family and $31/31$
for SuiteSparse) at every tested step count. The only non-convergent run was Nystr\"om with a $1\%$ sketch
on the SuiteSparse matrix \texttt{Cylshell/s1rmt3m1}.
These averages of absolute quantities differ from the
averages of per-instance ratios shown in
Figure~\ref{fig:rapid-condition-relative}.}
\label{tab:absolute-results}
\footnotesize
\setlength{\tabcolsep}{5pt}
\begin{tabular}{lrrrrrrr}
\toprule
& 
& \multicolumn{6}{c}{RAPID steps $T$} \\
\cmidrule(lr){3-8}
Dataset & {Jacobi baseline} & 10 & 20 & 50 & 100 & 200 & 500 \\
\midrule
\multicolumn{8}{l}{\textit{Condition number}} \\
Chebyshev
& 10,155.43 & 10,402.86 & 9,371.71 & 6,106.86 & 2,746.62 & 480.23 & 4.89 \\
Linear
& 10,007.98 & 9,342.59 & 8,066.53 & 4,816.50 & 1,857.98 & 228.63 & 4.82 \\
Logarithmic
& 9,978.65 & 7,429.91 & 5,697.62 & 2,983.64 & 1,316.00 & 330.48 & 6.07 \\
Power-law
& 10,010.99 & 8,940.11 & 7,707.07 & 4,845.05 & 2,216.73 & 425.81 & 4.94 \\
SuiteSparse
& 94,992.34 & 82,687.66 & 73,149.07 & 53,286.53 & 35,442.72 & 18,391.60 & 4,600.00 \\
\midrule
\multicolumn{8}{l}{\textit{CG iterations to target accuracy}} \\
Chebyshev
& 319.10 & 322.60 & 308.00 & 252.90 & 176.60 & 80.50 & 10.00 \\
Linear
& 127.90 & 122.00 & 115.50 & 90.30 & 59.70 & 29.80 & 10.00 \\
Logarithmic
& 373.30 & 322.70 & 283.40 & 207.80 & 140.70 & 74.50 & 17.80 \\
Power-law
& 332.80 & 313.80 & 291.60 & 233.60 & 163.20 & 79.00 & 10.00 \\
SuiteSparse
& 266.45 & 250.65 & 230.32 & 189.35 & 145.48 & 99.35 & 40.65 \\
\bottomrule
\end{tabular}
\end{table*}

%% file: paper/analysis_appx4.tex
\clearpage
\section{Proof of the Haar Block-Row-Min Guarantee}
\label{app:haar-block-row-min-proof}

This appendix proves Theorem~\ref{thm:maintext_conditioning_haar}. Let us define the block row selector:

\paragraph{Block row-min selector.}
For a partition \(\mathcal B\) of \([d]\) into blocks of size \(b\), 
Define
\begin{equation} \label{eq:block-row-min}
\operatorname{BRM}_{\mathcal B}(M)_{ij}
=
\mathbf 1\!\left\{
j=\argmin_{\ell\in B(i)\setminus\{i\}} M_{i\ell}
\right\},
\end{equation}
where \(B(i)\in\mathcal B\) is the block containing \(i\). Ties are broken
deterministically.
\begin{algorithm}[t]
\caption{Block row-min preconditioning}
\label{alg:preconditioner}
\begin{algorithmic}[1]
\State \textbf{Input:} \(A_0\succ0\), step size \(\alpha\), iterations \(T\), block size \(b\), mixer distribution \(\mathcal Q\).
\For{\(t=0,\ldots,T-1\)}
    \State Draw \(Q_t\sim\mathcal Q\) and a block partition \(\mathcal B_t\) of \([d]\) with block size \(b\).
    \State \(\widetilde A_t \gets Q_t^\top A_t Q_t\).
    \State \(R_t\gets \operatorname{BRM}_{\mathcal B_t}(\widetilde A_t)\).
    \State \(A_{t+1}\gets (I+\alpha R_t)^\top \widetilde A_t(I+\alpha R_t)\).
\EndFor
\State \Return \(A_T\).
\end{algorithmic}
\end{algorithm}
The theory uses $\bm Q_t\sim\operatorname{Haar}(O(d))$; experiments also use block-local choices based on product of Hadamard matrices of $\bm Q_t$.

Let us state the theorem in full generality here. We analyze the Gram recursion extracted in Section~\ref{sec:setup-extracted-preconditioner}:
\begin{equation}
    \widetilde A_t=Q_t^\top A_tQ_t,
    \qquad
    A_{t+1}
    =
    (I+\alpha R_t)^\top\widetilde A_t(I+\alpha R_t),
    \qquad
    R_t=\operatorname{BRM}_{\mathcal B_t}(\widetilde A_t),
    \label{eq:theory-recursion}
\end{equation}
started from an arbitrary $A_0\succ0$.  The theorem is proved for fresh uniform random orthogonal mixing, $Q_t\sim\operatorname{Haar}(O(d))$.  This distributional assumption is used only in the local row-min drift calculation; the Taylor expansion, trace identities, block-load bound, spectral lower bound, and stopping-time argument hold for any orthogonal mixer.

For the block selector defined in \eqref{eq:block-row-min}, each row selects one off-diagonal entry from its block and each column can be selected by at most $b-1$ rows.  Thus
\begin{equation}
    \|R_t\|_{\rm op}^2\le b-1.
    \label{eq:theory-load-bound}
\end{equation}
This deterministic load bound is the only property of the block mask used in the Taylor remainder.

\begin{theorem}[Uniform-random-rotation block row-min preconditioning]
\label{thm:haar-block-row-min}
Let $d\ge3$, $3\le b\le d$, and $b\mid d$.  Fix a target condition number $\kappa_\star>1$, and define
\begin{equation}
\rho_{\kappa_\star}
=
\max\left\{
1-\frac{1-\kappa_\star^{-1}}4,
\left(1+\frac{\kappa_\star-1}{4}\right)^{-1}
\right\}<1.
\end{equation}
Choose
\begin{equation}
    \beta\ge \frac{\log(8d)}{-\log\rho_{\kappa_\star}}.
\end{equation}
Let
\begin{equation}
    \nu_{b,d}
    =
    -\sqrt{d-1}\,\E\left[\min_{1\le r\le b-1}U_r\right],
    \qquad
    U\sim\operatorname{Unif}(S^{d-2}).
\end{equation}
There are universal constants $C_{\rm step},C_{\rm hit}>0$ such that, for
\begin{equation}
    \alpha
    =
    \frac{\nu_{b,d}}
    {C_{\rm step}\,\beta(b-1)\sqrt{d-1}},
\end{equation}
the iteration \eqref{eq:theory-recursion} with fresh $Q_t\sim\operatorname{Haar}(O(d))$ satisfies
\begin{equation}
    \E[\tau_{\kappa_\star}]
    \le
    C_{\rm hit}
    \frac{\beta(b-1)(d-1)}{\nu_{b,d}^2}
    \log\kappa(A_0),
    \qquad
    \tau_{\kappa_\star}=\inf\{t:\kappa(A_t)\le\kappa_\star\}.
\end{equation}
For each fixed $b\ge3$, $\nu_{b,d}^2=\Theta_b(1)$ uniformly in $d$.  Hence, for fixed $b$ and fixed $\kappa_\star$,
\begin{equation}
    \E[\tau_{\kappa_\star}]
    =
    O_{\kappa_\star,b}\left(d\log d\,\log\kappa(A_0)\right).
\end{equation}
\end{theorem}

Theorem~\ref{thm:haar-block-row-min} is a refinement of  Theorem~\ref{thm:maintext_conditioning_haar} from the main text. The rest of the analysis will focus on proving this result. In the process, we will show improved versions of  Lemma~\ref{lemma:maintext_drift} and Lemma~\ref{lemma:maintext_taylor}.

\subsection{Setup and potential}

Let \(A_0\succ0\) and let
\[
A_{t+1}
=
(I+\alpha R_t)^\top \widetilde A_t(I+\alpha R_t),
\qquad
\widetilde A_t=Q_t^\top A_tQ_t,
\qquad
Q_t\sim\Haar(O(d)).
\]
Fix a block partition \([d]=B_1\sqcup\cdots\sqcup B_{d/b}\), \(|B_r|=b\).  For \(i\in B_r\), define
\[
\tau_i=\argmin_{j\in B_r,\ j\neq i}(\widetilde A_t)_{ij},
\qquad
(R_t)_{ij}=\1\{j=\tau_i\},
\]
with deterministic tie-breaking.  A fresh independent random partition may also be used, but the Haar proof only needs the candidate sets to be fixed independently of \(Q_t\).

For \(\beta\ge1\), define
\[
\Phi_\beta(A)
=
\frac1\beta\log\frac{\tr(A^\beta)}d
+
\frac1\beta\log\frac{\tr(A^{-\beta})}d.
\]
This potential is scale-invariant and satisfies
\[
0\le \Phi_\beta(A)\le \log\kappa(A) \le \Phi_\beta(A) + \frac{2 \log d}{\beta}.
\]
Let \(\lambda=(\lambda_1,\ldots,\lambda_d)\) be the eigenvalues of \(A\), and let
\[
p\sim\Dir(1/2,\ldots,1/2).
\]
Define
\[
\Var_p(\lambda)
=
\sum_i p_i\lambda_i^2-
\left(\sum_i p_i\lambda_i\right)^2,
\]
\[
\Cov_p(\lambda,\lambda^\gamma)
=
\sum_i p_i\lambda_i^{\gamma+1}
-
\left(\sum_i p_i\lambda_i\right)
\left(\sum_i p_i\lambda_i^\gamma\right).
\]
Ratios below are defined to be zero when \(\Var_p(\lambda)=0\).  Set
\[
s_\beta(A)
=
\E_p\left[
\frac{\Cov_p(\lambda,\lambda^\beta)}{\sqrt{\Var_p(\lambda)}}
\right],
\qquad
s_{-\beta}(A)
=
-
\E_p\left[
\frac{\Cov_p(\lambda,\lambda^{-\beta})}{\sqrt{\Var_p(\lambda)}}
\right],
\]
and
\begin{equation}\label{eq:Sbeta}
    S_\beta(A)
=
\frac{d\,s_\beta(A)}{\tr(A^\beta)}
+
\frac{d\,s_{-\beta}(A)}{\tr(A^{-\beta})}.
\end{equation}

\subsection{Block masking controls the load}

For each row \(i\), exactly one entry of \(R_t\) is nonzero, so \(\|R_t\|_\infty=1\).  Fix a column \(j\in B_r\).  Only rows in \(B_r\setminus\{j\}\) can select column \(j\), so the column has at most \(b-1\) nonzeros.  Therefore
\[
\|R_t\|_1\le b-1,
\qquad
\|R_t\|_{\rm op}
\le
\sqrt{\|R_t\|_1\|R_t\|_\infty}
\le
\sqrt{b-1}.
\]

\subsection{Taylor drift estimate}

\begin{lemma}[Taylor estimate]
\label{lem:taylor-estimate}
Let \(\widetilde A\succ0\), \(R\in\mathbb{R}^{d\times d}\), and
\[
A(s)=(I+sR)^\top\widetilde A(I+sR),
\qquad
f(s)=\Phi_\beta(A(s)).
\]
If \(\beta\ge1\) and \(\alpha\|R\|_{\rm op}\le1/2\), then
\[
f'(0)
=
2\left(
\frac{\tr(R\widetilde A^\beta)}{\tr(\widetilde A^\beta)}
-
\frac{\tr(R\widetilde A^{-\beta})}{\tr(\widetilde A^{-\beta})}
\right),
\]
and there is a universal constant \(C_{\rm Tay}\) such that, for every \(s\in[0,\alpha]\),
\[
f''(s)\le C_{\rm Tay}\beta\|R\|_{\rm op}^2.
\]
One may take \(C_{\rm Tay}=40\).
\end{lemma}

\begin{proof}
Since
\[
A'(0)=R^\top\widetilde A+\widetilde A R,
\]
the trace-power chain rule gives
\[
\frac{d}{ds}\bigg|_{s=0}\frac1\beta\log\tr(A(s)^\beta)
=
\frac{\tr(\widetilde A^{\beta-1}(R^\top\widetilde A+\widetilde A R))}
{\tr(\widetilde A^\beta)}.
\]
By cyclicity of trace,
\[
\tr(\widetilde A^{\beta-1}(R^\top\widetilde A+\widetilde A R))
=2\tr(R\widetilde A^\beta).
\]
The negative-power term similarly gives
\[
\frac{d}{ds}\bigg|_{s=0}\frac1\beta\log\tr(A(s)^{-\beta})
=
-2\frac{\tr(R\widetilde A^{-\beta})}{\tr(\widetilde A^{-\beta})}.
\]
This proves the formula for \(f'(0)\).

For the curvature bound, put
\[
P_s=I+sR,
\qquad
A_s=P_s^\top\widetilde A P_s,
\qquad
C_s=P_s^{-1}R.
\]
For \(s\in[0,\alpha]\), the assumption \(\alpha\|R\|_{\rm op}\le1/2\) implies
\[
\|P_s^{-1}\|_{\rm op}\le2,
\qquad
\|C_s\|_{\rm op}\le2\|R\|_{\rm op}.
\]
Moreover,
\[
A(s+t)=(I+tC_s)^\top A_s(I+tC_s).
\]

Let \(p=2\beta\ge2\).  For the Schatten \(p\)-norm, the standard 2-smoothness estimate for \(N(X)=\|X\|_p^2\) gives
\[
DN(X)[Y]\le2\|X\|_p\|Y\|_p,
\qquad
D^2N(X)[Y,Y]\le2(p-1)\|Y\|_p^2.
\]
Consequently, for a twice differentiable path \(X(t)\),
\[
\frac{d^2}{dt^2}\log\|X(t)\|_p^2
\le
2(p-1)\frac{\|X'(t)\|_p^2}{\|X(t)\|_p^2}
+
2\frac{\|X''(t)\|_p}{\|X(t)\|_p}.
\]

For the positive-power term, write
\[
\frac1\beta\log\tr(A(s+t)^\beta)
=
\log\|A_s^{1/2}(I+tC_s)\|_{2\beta}^2.
\]
With \(X_+(t)=A_s^{1/2}(I+tC_s)\), we have \(X_+''(0)=0\), and Schatten submultiplicativity gives
\[
\frac{d^2}{dt^2}\bigg|_{t=0}
\frac1\beta\log\tr(A(s+t)^\beta)
\le
(4\beta-2)\|C_s\|_{\rm op}^2.
\]
For the negative-power term,
\[
A(s+t)^{-1}=(I+tC_s)^{-1}A_s^{-1}(I+tC_s)^{-\top},
\]
so
\[
\frac1\beta\log\tr(A(s+t)^{-\beta})
=
\log\|(I+tC_s)^{-1}A_s^{-1/2}\|_{2\beta}^2.
\]
With \(X_-(t)=(I+tC_s)^{-1}A_s^{-1/2}\),
\[
X_-'(0)=-C_sA_s^{-1/2},
\qquad
X_-''(0)=2C_s^2A_s^{-1/2},
\]
and hence
\[
\frac{d^2}{dt^2}\bigg|_{t=0}
\frac1\beta\log\tr(A(s+t)^{-\beta})
\le
6\beta\|C_s\|_{\rm op}^2.
\]
Combining the two terms and using \(\|C_s\|_{\rm op}\le2\|R\|_{\rm op}\) yields
\[
f''(s)\le40\beta\|R\|_{\rm op}^2.
\]
\end{proof}

Applying Lemma~\ref{lem:taylor-estimate} to one update and using \(\|R\|_{\rm op}^2\le b-1\) gives
\[
\Phi_\beta(A^+)
\le
\Phi_\beta(A)
+
2\alpha\left(
\frac{\tr(R\widetilde A^\beta)}{\tr(A^\beta)}
-
\frac{\tr(R\widetilde A^{-\beta})}{\tr(A^{-\beta})}
\right)
+
C_{\rm Tay}\alpha^2\beta(b-1),
\]
where \(\tr(\widetilde A^{\pm\beta})=\tr(A^{\pm\beta})\) by orthogonal invariance.

\subsection{Block-local row-min drift}

\begin{lemma}[Haar row-min drift]
\label{lem:haar-row-min-drift}
For fixed \(A\succ0\), let \(Q\sim\Haar(O(d))\), \(\widetilde A=Q^\top A Q\), and let \(R\) be the block row-min selector.  Then
\[
\E\left[
\frac{\tr(R\widetilde A^\beta)}{\tr(A^\beta)}
-
\frac{\tr(R\widetilde A^{-\beta})}{\tr(A^{-\beta})}
\,\middle|\,A
\right]
=
-
\frac{\nu_{b,d}}{\sqrt{d-1}}S_\beta(A).
\]
\end{lemma}

\begin{proof}
For any real \(\gamma\), symmetry of \(\widetilde A^\gamma\) gives
\[
(R\widetilde A^\gamma)_{ii}
=
\sum_jR_{ij}(\widetilde A^\gamma)_{ji}
=
(\widetilde A^\gamma)_{\tau_i i}
=
(\widetilde A^\gamma)_{i\tau_i}.
\]
Thus
\[
\tr(R\widetilde A^\gamma)=\sum_{i=1}^d(\widetilde A^\gamma)_{i\tau_i}.
\]
It suffices to compute the one-row expectation.  For each row, the candidate set has size \(b-1\) and is independent of \(Q\).  By exchangeability of Haar coordinates, the one-row law depends only on this size.  We analyze row \(1\) with a fixed candidate set \(J\subset[d]\setminus\{1\}\), \(|J|=b-1\).

Let
\[
x_j=(\widetilde A)_{1j},
\qquad
(y_\gamma)_j=(\widetilde A^\gamma)_{1j},
\qquad j\neq1,
\]
and
\[
\tau=\argmin_{j\in J}x_j.
\]
Then the row contribution is \((y_\gamma)_\tau\).

Write \(A=V\Lambda V^\top\) and set \(U=Q^\top V\).  Then \(U\sim\Haar(O(d))\).  Conditional on the first row \(u\in S^{d-1}\) of \(U\), choose an orthonormal basis \(B\in\mathbb{R}^{d\times(d-1)}\) for \(u^\perp\).  The remaining rows of \(U\) form a Haar frame in \(u^\perp\), so there is \(W\sim\Haar(O(d-1))\) such that
\[
x=aW^\top,
\qquad
 y_\gamma=c_\gamma W^\top,
\qquad
 a=u\Lambda B,
\qquad
 c_\gamma=u\Lambda^\gamma B.
\]
The vectors \(a,c_\gamma\) are deterministic after conditioning on \(u\).

Assume first that \(\|a\|>0\).  Decompose
\[
c_\gamma
=
\frac{\langle a,c_\gamma\rangle}{\|a\|^2}a+c_\gamma^\perp,
\qquad
c_\gamma^\perp\perp a.
\]
Then
\[
y_\gamma
=
\frac{\langle a,c_\gamma\rangle}{\|a\|^2}x+c_\gamma^\perp W^\top.
\]
The index \(\tau\) is a function only of \(x=aW^\top\).  Conditional on \(x\), the law of \(W\) is invariant under rotations fixing \(a\).  These rotations leave \(x\), and hence \(\tau\), unchanged, while they average the \(a^\perp\)-component to zero.  Therefore
\[
\E[(c_\gamma^\perp W^\top)_\tau\mid a,c_\gamma,J]=0,
\]
and
\[
\E[(y_\gamma)_\tau\mid a,c_\gamma,J]
=
\frac{\langle a,c_\gamma\rangle}{\|a\|^2}
\E[x_\tau\mid a,J].
\]
If \(\|a\|=0\), both sides are interpreted using the zero-ratio convention.

Since \(x=aW^\top\stackrel d=\|a\|U\), with \(U\sim\Unif(S^{d-2})\),
\[
\E[x_\tau\mid a,J]
=
-\frac{\nu_{b,d}}{\sqrt{d-1}}\|a\|.
\]
Thus
\[
\E[(y_\gamma)_\tau\mid a,c_\gamma,J]
=
-
\frac{\nu_{b,d}}{\sqrt{d-1}}
\frac{\langle a,c_\gamma\rangle}{\|a\|}.
\]

Now average over \(u\).  If \(p_i=u_i^2\), then \(p\sim\Dir(1/2,\ldots,1/2)\).  Since \(BB^\top=I-u^\top u\),
\[
\|a\|^2
=
u_{\rm var}
:=
\sum_i p_i\lambda_i^2
-
\left(\sum_i p_i\lambda_i\right)^2
=
\Var_p(\lambda),
\]
\[
\langle a,c_\beta\rangle
=
\Cov_p(\lambda,\lambda^\beta),
\qquad
\langle a,c_{-\beta}\rangle
=
\Cov_p(\lambda,\lambda^{-\beta}).
\]
Therefore
\[
\E[(y_\beta)_\tau\mid A]
=
-
\frac{\nu_{b,d}}{\sqrt{d-1}}s_\beta(A),
\]
and
\[
\E[(y_{-\beta})_\tau\mid A]
=
\frac{\nu_{b,d}}{\sqrt{d-1}}s_{-\beta}(A).
\]
Summing over the \(d\) rows gives
\[
\E[\tr(R\widetilde A^\beta)\mid A]
=
-
\frac{\nu_{b,d}}{\sqrt{d-1}}d\,s_\beta(A),
\]
\[
\E[\tr(R\widetilde A^{-\beta})\mid A]
=
\frac{\nu_{b,d}}{\sqrt{d-1}}d\,s_{-\beta}(A).
\]
Dividing by the deterministic traces yields the claimed identity.
\end{proof}

\subsection{Lower bound on the spectral factor}

\begin{lemma}[Spectral lower bound]
\label{lem:spectral-lower-bound}
Let \(\kappa_\star>1\), and define
\[
\rho_{\kappa_\star}
=
\max\left\{
1-\frac{1-\kappa_\star^{-1}}4,
\left(1+\frac{\kappa_\star-1}{4}\right)^{-1}
\right\}.
\]
If \(\kappa(A)\ge\kappa_\star\) and
\[
\beta\ge\frac{\log(8d)}{-\log\rho_{\kappa_\star}},
\]
then
\[
S_\beta(A)\ge\frac14.
\]
\end{lemma}

\begin{proof}
By scale invariance, set \(\lambda_{\min}=1\), \(\lambda_{\max}=\kappa\ge\kappa_\star\), and \(\Delta=\kappa-1\).  Define
\[
\mu_+
=
\frac{\sum_i\lambda_i^{\beta+1}}{\sum_i\lambda_i^\beta},
\qquad
\mu_-
=
\frac{\sum_i\lambda_i^{1-\beta}}{\sum_i\lambda_i^{-\beta}},
\qquad
\bar\lambda=\frac1d\sum_i\lambda_i.
\]
Since \(\lambda_i\in[1,\kappa]\),
\[
\Var_p(\lambda)\le\frac{\Delta^2}{4}.
\]
Monotonicity gives \(\Cov_p(\lambda,\lambda^\beta)\ge0\) and \(-\Cov_p(\lambda,\lambda^{-\beta})\ge0\).  Hence
\[
s_\beta
\ge
\frac2\Delta\E_p[\Cov_p(\lambda,\lambda^\beta)],
\qquad
s_{-\beta}
\ge
\frac2\Delta\E_p[-\Cov_p(\lambda,\lambda^{-\beta})].
\]
For \(p\sim\Dir(1/2,\ldots,1/2)\),
\[
\E_p[\Cov_p(u,v)]
=
\frac{d\sum_i u_i v_i-(\sum_i u_i)(\sum_i v_i)}{d(d+2)}.
\]
Applying this identity gives
\[
\E_p[\Cov_p(\lambda,\lambda^\beta)]
=
\frac{\tr(A^\beta)}{d+2}(\mu_+-\bar\lambda),
\]
\[
\E_p[-\Cov_p(\lambda,\lambda^{-\beta})]
=
\frac{\tr(A^{-\beta})}{d+2}(\bar\lambda-\mu_-).
\]
Therefore
\[
S_\beta(A)
\ge
\frac{2d}{\Delta(d+2)}(\mu_+-\mu_-)
\ge
\frac{\mu_+-\mu_-}{\Delta}.
\]

It remains to show \(\mu_+-\mu_-\ge\Delta/4\).  Let
\[
r=1+\frac\Delta4,
\qquad
s=\kappa-\frac\Delta4.
\]
Then \(s-r=\Delta/2\).  Let
\[
W_+=
\frac{\sum_{\lambda_i<s}\lambda_i^\beta}{\sum_i\lambda_i^\beta}.
\]
Since at least one eigenvalue equals \(\kappa\),
\[
W_+\le d\left(\frac{s}{\kappa}\right)^\beta
\le d\rho_{\kappa_\star}^\beta
\le \frac18.
\]
Therefore
\[
\mu_+\ge s-(s-1)W_+
\ge s-\frac{3\Delta}{32}.
\]
Similarly, with
\[
W_-=
\frac{\sum_{\lambda_i>r}\lambda_i^{-\beta}}{\sum_i\lambda_i^{-\beta}},
\]
one has
\[
W_-\le d r^{-\beta}
\le d\rho_{\kappa_\star}^\beta
\le \frac18,
\]
and hence
\[
\mu_-\le r+(\kappa-r)W_-
\le r+\frac{3\Delta}{32}.
\]
Thus
\[
\mu_+-\mu_-
\ge
(s-r)-\frac{6\Delta}{32}
=
\frac{5\Delta}{16}
\ge
\frac\Delta4.
\]
The lemma follows.
\end{proof}

\subsection{Hitting-time bound}

For \(t<\tau_{\kappa_\star}\), Lemma~\ref{lem:spectral-lower-bound} gives \(S_\beta(A_t)\ge1/4\).  Combining Lemma~\ref{lem:haar-row-min-drift} with the Taylor estimate gives
\[
\E[\Phi_\beta(A_{t+1})-\Phi_\beta(A_t)\mid A_t]
\le
-\frac{\alpha\nu_{b,d}}{2\sqrt{d-1}}
+
C_{\rm Tay}\alpha^2\beta(b-1).
\]
Choose
\[
\alpha
=
\frac{\nu_{b,d}}{8C_{\rm Tay}\beta(b-1)\sqrt{d-1}},
\]
and enlarge the universal constant in Theorem~\ref{thm:haar-block-row-min}, if necessary, to ensure \(\alpha\|R_t\|_{\rm op}\le1/2\).  Then
\[
\E[\Phi_\beta(A_{t+1})-\Phi_\beta(A_t)\mid A_t]
\le
-\delta,
\qquad
\delta
=
\frac{3\nu_{b,d}^2}{64C_{\rm Tay}\beta(b-1)(d-1)}.
\]
Therefore
\[
\Phi_\beta(A_{t\wedge\tau_{\kappa_\star}})
+
\delta(t\wedge\tau_{\kappa_\star})
\]
is a supermartingale.  Since \(\Phi_\beta\ge0\), optional stopping gives
\[
\E[t\wedge\tau_{\kappa_\star}]
\le
\frac{\Phi_\beta(A_0)}{\delta}.
\]
Letting \(t\to\infty\) and using \(\Phi_\beta(A_0)\le\log\kappa(A_0)\) yields
\[
\E[\tau_{\kappa_\star}]
\le
C
\frac{\beta(b-1)(d-1)}{\nu_{b,d}^2}
\log\kappa(A_0).
\]

It remains to record the fixed-block behavior of \(\nu_{b,d}\).  Since \(b-1\ge2\),
\[
-\E\min_{1\le r\le b-1}U_r
\ge
-\E\min(U_1,U_2)
=
\frac12\E|U_1-U_2|.
\]
By rotational invariance, \(U_1-U_2\stackrel d=\sqrt2\,U_1\), so
\[
\nu_{b,d}
\ge
\sqrt{\frac{d-1}{2}}\,\E|U_1|.
\]
For \(U\sim\Unif(S^{d-2})\),
\[
\E|U_1|
=
\frac{\Gamma((d-1)/2)}{\sqrt\pi\,\Gamma(d/2)},
\]
which gives a positive lower bound uniformly in \(d\).  Conversely,
\[
-\min_{1\le r\le b-1}U_r
\le
\max_{1\le r\le b-1}|U_r|
\le
\left(\sum_{r=1}^{b-1}U_r^2\right)^{1/2}.
\]
Thus, by Jensen,
\[
\nu_{b,d}
\le
\sqrt{d-1}\left(\E\sum_{r=1}^{b-1}U_r^2\right)^{1/2}
=
\sqrt{b-1}.
\]
Hence \(\nu_{b,d}^2=\Theta_b(1)\) for fixed \(b\).  Since \(\beta=O_{\kappa_\star}(\log d)\), Theorem~\ref{thm:haar-block-row-min} follows.

\subsection{Fast structured mixing and the required transfer condition}
\label{subsec:fast-hadamard-condition}

Dense Haar mixing is the clean object for analysis, but it is not the intended implementation for large matrices.  Sampling and applying a dense Haar matrix costs cubic time at the scale of a dense \(d\times d\) Gram matrix.  A much faster alternative is to use a product of normalized Walsh-Hadamard transforms and random sign diagonals.  Let \(\mathsf W_d\) denote the normalized Walsh-Hadamard matrix, and let \(D_1,\ldots,D_k\) be independent diagonal matrices with Rademacher entries.  A typical fast mixer is
\[
Q_{\rm FHT}^\top
=
\Pi\,\mathsf W_dD_k\mathsf W_dD_{k-1}\cdots \mathsf W_dD_1,
\]
where \(\Pi\) is an independent random permutation.  Each factor is orthogonal, and multiplication by \(\mathsf W_d\) can be performed with the fast Hadamard transform.  Thus forming
\(
Q_{\rm FHT}^\top A Q_{\rm FHT}
\)
for dense \(A\) costs \(O(kd^2\log d)\), rather than the dense-rotation cost.  The sparse row-min update then costs \(O(d^2)\), and the block row-min search costs \(O(db)\) once the relevant entries of \(\widetilde A\) are available.

The theorem above does not automatically apply to \(Q_{\rm FHT}\).  Haar mixing supplies exact conditional rotational invariance, which is used in the row-min drift lemma.  A finite product of Hadamard/sign factors is a structured discrete distribution and need not have that exact invariance.  A single Hadamard-sign layer is certainly insufficient in the worst case: if \(Q=\mathsf W_dD\) and \(A=\mathsf W_d\Lambda\mathsf W_d^\top\), then \(Q^\top A Q=D\Lambda D=\Lambda\), so the mixed matrix is diagonal and the row-min signal vanishes.  Multiple layers and a random permutation are used to avoid such alignment effects. This worst-case observation does not contradict the empirical effectiveness of a single mixing stage per RAPID iteration ($k=1$), since successive iterations introduce fresh mixing interleaved with sparse corrections, allowing the effects of mixing to accumulate over time.

The rest of the proof, however, is insensitive to the source of orthogonal mixing.  The Taylor expansion, trace invariance, load bound, spectral lower bound, and stopping-time argument all hold for any orthogonal distribution.  The only missing ingredient is a local row-min comparison estimate.  We state it explicitly.

For an orthogonal mixing distribution \(\mathcal Q\), define, for a fixed row \(i\), block candidate set \(J_i\), and exponent \(\gamma\in\{\beta,-\beta\}\),
\[
x_j=(Q^\top A Q)_{ij},
\qquad
 y_j^{(\gamma)}=((Q^\top A Q)^\gamma)_{ij},
\qquad j\in J_i,
\]
and
\[
\tau_i=\argmin_{j\in J_i}x_j.
\]
Let
\[
F_{i,J_i}^{(\gamma)}(Q,A)
:=
 y_{\tau_i}^{(\gamma)}.
\]
Define the local Haar-comparison errors
\[
\delta_{\mathcal Q,\gamma}(A)
:=
\max_i
\left|
\E_{Q\sim\mathcal Q}F_{i,J_i}^{(\gamma)}(Q,A)
-
\E_{Q\sim\Haar}F_{i,J_i}^{(\gamma)}(Q,A)
\right|,
\]
and the normalized error
\[
\varepsilon_{\mathcal Q,\beta}(A)
:=
\frac{d\,\delta_{\mathcal Q,\beta}(A)}{\tr(A^\beta)}
+
\frac{d\,\delta_{\mathcal Q,-\beta}(A)}{\tr(A^{-\beta})}.
\]

\begin{proposition}[Transfer from Haar to a structured orthogonal mixer]
\label{prop:orthogonal-transfer}
Let \(\mathcal Q\) be any distribution on \(O(d)\).  Suppose that for every \(A\succ0\) with \(\kappa(A)\ge\kappa_\star\),
\[
\varepsilon_{\mathcal Q,\beta}(A)
\le
\frac{\nu_{b,d}}{8\sqrt{d-1}}.
\]
Then the block row-min iteration with \(Q_t\sim\mathcal Q\) satisfies the same hitting-time guarantee as Theorem~\ref{thm:haar-block-row-min}, up to universal constant factors:
\[
\E[\tau_{\kappa_\star}]
\le
C
\frac{\beta(b-1)(d-1)}{\nu_{b,d}^2}
\log\kappa(A_0).
\]
\end{proposition}

\begin{proof}
The Haar drift identity gives
\[
\mathcal D_{\rm Haar}(A)
:=
\E_{\Haar}\left[
\frac{\tr(R\widetilde A^\beta)}{\tr(A^\beta)}
-
\frac{\tr(R\widetilde A^{-\beta})}{\tr(A^{-\beta})}
\,\middle|\,A
\right]
=
-
\frac{\nu_{b,d}}{\sqrt{d-1}}S_\beta(A).
\]
By the definition of \(\varepsilon_{\mathcal Q,\beta}\), the corresponding drift under \(Q\sim\mathcal Q\) satisfies
\[
\mathcal D_{\mathcal Q}(A)
\le
\mathcal D_{\rm Haar}(A)+\varepsilon_{\mathcal Q,\beta}(A).
\]
If \(\kappa(A)\ge\kappa_\star\), then \(S_\beta(A)\ge1/4\), and the assumed comparison bound gives
\[
\mathcal D_{\mathcal Q}(A)
\le
-
\frac{\nu_{b,d}}{4\sqrt{d-1}}
+
\frac{\nu_{b,d}}{8\sqrt{d-1}}
=
-
\frac{\nu_{b,d}}{8\sqrt{d-1}}.
\]
Substituting this into the same Taylor estimate used in the Haar proof yields
\[
\E[\Phi_\beta(A_{t+1})-\Phi_\beta(A_t)\mid A_t]
\le
-
\frac{\alpha\nu_{b,d}}{4\sqrt{d-1}}
+
C_{\rm Tay}\alpha^2\beta(b-1)
\]
until \(\tau_{\kappa_\star}\).  Choosing
\(
\alpha\asymp \nu_{b,d}/(\beta(b-1)\sqrt{d-1})
\)
and applying the same optional-stopping argument proves the claim.
\end{proof}

Proposition~\ref{prop:orthogonal-transfer} is stated as a local comparison condition, not as a new convergence theorem for fast Hadamard products.  Proving the condition for
\(
Q_{\rm FHT}
\)
requires showing that the hard row-min statistic
\(
F_{i,J_i}^{(\gamma)}
\)
has nearly the same expectation under the structured mixer as under Haar.  This is the precise theoretical gap left by replacing continuous Haar rotations with fast structured orthogonal transforms.  Existing incoherence and fast Johnson-Lindenstrauss results for randomized Hadamard transforms suggest why such a comparison should be plausible after several independent sign/Hadamard layers, but they do not by themselves supply the conditional rotational invariance used in the Haar proof.  Empirically, we therefore use the fast Hadamard mixer for scalability, while the theorem above gives the exact guarantee for Haar mixing and identifies the comparison estimate needed to transfer the guarantee to the fast implementation.

\subsection{Spiky spectra give a \texorpdfstring{\(\sqrt d\)}{sqrt(d)} drift gain}
\label{app:spiky}
In the worst-case argument above we only used the uniform lower bound
\[
S_\beta(A)\ge c_{\kappa_\star}>0
\qquad
\text{whenever}
\qquad
\kappa(A)\ge \kappa_\star.
\]
For spiky spectra, \(S_\beta(A)\) can be much larger. We record the
simplest case, where the spectrum consists of one large spike and a flat
bulk.

\begin{lemma}[Rank-one spike]
Assume, after rescaling, that
\[
\lambda_1=\kappa,
\qquad
\lambda_2=\cdots=\lambda_d=1,
\qquad
\kappa>1.
\]
Let
\[
M=\kappa^\beta.
\]
Then
\[
S_\beta(A)
=
d A_d
\left[
\frac{M-1}{M+d-1}
+
\frac{1-M^{-1}}{M^{-1}+d-1}
\right],
\]
where
\[
A_d
=
\E\sqrt{q(1-q)},
\qquad
q\sim \mathrm{Beta}\left(\frac12,\frac{d-1}{2}\right).
\]
Moreover,
\[
A_d
=
\frac{2}{d}
\frac{\Gamma(d/2)}
{\sqrt\pi\,\Gamma((d-1)/2)}
=
\Theta(d^{-1/2}).
\]
Consequently, if
\[
\kappa^\beta\ge 2d,
\]
then
\[
S_\beta(A)\ge c\sqrt d
\]
for a universal numerical constant \(c>0\).
In fact, in this regime,
\[
S_\beta(A)=\Theta(\sqrt d).
\]
\end{lemma}

\begin{proof}
Let
\[
p\sim \mathrm{Dir}\left(\frac12,\ldots,\frac12\right),
\qquad
q=p_1.
\]
For the rank-one spike,
\[
\lambda_i = 1+(\kappa-1)\mathbf 1_{\{i=1\}}.
\]
Set
\[
\Delta=\kappa-1.
\]
Then
\[
\lambda^\beta_i
=
1+(M-1)\mathbf 1_{\{i=1\}},
\qquad
\lambda^{-\beta}_i
=
1+(M^{-1}-1)\mathbf 1_{\{i=1\}}.
\]
A direct calculation gives
\[
\Var_p(\lambda)
=
\Delta^2 q(1-q),
\]
\[
\Cov_p(\lambda,\lambda^\beta)
=
\Delta(M-1)q(1-q),
\]
\[
-\Cov_p(\lambda,\lambda^{-\beta})
=
\Delta(1-M^{-1})q(1-q).
\]
Therefore, using the zero-ratio convention when \(\Var_p(\lambda)=0\),
\[
s_\beta(A)
=
(M-1)\E\sqrt{q(1-q)},
\]
\[
s_{-\beta}(A)
=
(1-M^{-1})\E\sqrt{q(1-q)}.
\]
Writing
\[
A_d:=\E\sqrt{q(1-q)},
\]
we have
\[
s_\beta(A)=(M-1)A_d,
\qquad
s_{-\beta}(A)=(1-M^{-1})A_d.
\]
Also,
\[
\tr(A^\beta)=M+d-1,
\qquad
\tr(A^{-\beta})=M^{-1}+d-1.
\]
Hence
\[
S_\beta(A)
=
\frac{d\,s_\beta(A)}{\tr(A^\beta)}
+
\frac{d\,s_{-\beta}(A)}{\tr(A^{-\beta})}
\]
equals
\[
d A_d
\left[
\frac{M-1}{M+d-1}
+
\frac{1-M^{-1}}{M^{-1}+d-1}
\right].
\]

It remains to estimate \(A_d\). Since
\[
q\sim \mathrm{Beta}\left(\frac12,\frac{d-1}{2}\right),
\]
we get
\[
A_d
=
\frac{
B\left(1,d/2\right)
}{
B\left(1/2,(d-1)/2\right)
}.
\]
Using
\[
B\left(1,d/2\right)=\frac{2}{d}
\]
and
\[
B\left(1/2,(d-1)/2\right)
=
\frac{
\sqrt\pi\,\Gamma((d-1)/2)
}{
\Gamma(d/2)
},
\]
we obtain
\[
A_d
=
\frac{2}{d}
\frac{\Gamma(d/2)}
{\sqrt\pi\,\Gamma((d-1)/2)}.
\]
By the standard gamma-ratio estimate,
\[
\frac{\Gamma(d/2)}
{\Gamma((d-1)/2)}
=
\Theta(\sqrt d),
\]
and hence
\[
A_d=\Theta(d^{-1/2}).
\]
Thus
\[
dA_d=\Theta(\sqrt d).
\]

Finally, if \(M=\kappa^\beta\ge 2d\), then
\[
\frac{M-1}{M+d-1}
\ge c_0
\]
for a universal constant \(c_0>0\). The second term is nonnegative, so
\[
S_\beta(A)
\ge
c_0 dA_d
\ge
c\sqrt d.
\]
The upper bound \(S_\beta(A)\le C\sqrt d\) follows from the same formula
and the fact that both bracketed terms are at most \(1\).
\end{proof}

\begin{remark}[Interpretation]
The uniform theorem only uses \(S_\beta(A)\gtrsim 1\). For a rank-one
spike with \(\kappa^\beta\gtrsim d\), the useful spectral drift is larger:
\[
S_\beta(A)\asymp \sqrt d.
\]
This happens because a Haar row assigns mass
\[
q=p_1\sim \mathrm{Beta}\left(\frac12,\frac{d-1}{2}\right)
\]
to the spiked eigendirection, and
\[
\E\sqrt{q(1-q)}\asymp d^{-1/2}.
\]
The definition of \(S_\beta\) contains an additional factor of \(d\), so
the net scale is \(\sqrt d\).
\end{remark}

\subsubsection{Consequence for the drift rate in the spiky regime}

We now spell out the resulting improvement in the one-step drift. Recall
the general drift estimate
\[
\E[\Phi_\beta(A_{t+1})-\Phi_\beta(A_t)\mid A_t]
\le
-
\frac{2\alpha\nu_{b,d}}{\sqrt{d-1}}
S_\beta(A_t)
+
C_{\rm Tay}\alpha^2\beta(b-1).
\]
Suppose that, on some portion of the trajectory before reaching the
target condition number, the iterates satisfy the stronger spiky lower
bound
\[
S_\beta(A_t)\ge \sigma\sqrt d
\]
for some \(\sigma>0\). Using the same globally safe step size
\[
\alpha
=
\frac{\nu_{b,d}}
{8C_{\rm Tay}\beta(b-1)\sqrt{d-1}},
\]
we obtain
\[
\E[\Phi_\beta(A_{t+1})-\Phi_\beta(A_t)\mid A_t]
\le
-
c
\frac{\sigma\nu_{b,d}^2}
{C_{\rm Tay}\beta(b-1)\sqrt d}
\]
for a universal numerical constant \(c>0\), after absorbing the Taylor
remainder into the leading negative term.

Consequently, if
\[
S_\beta(A_t)\ge \sigma\sqrt d
\]
holds until the hitting time
\[
\tau_{\kappa_\star}
=
\inf\{t:\kappa(A_t)\le \kappa_\star\},
\]
then the stopped-supermartingale argument gives
\[
\E[\tau_{\kappa_\star}]
\le
C
\frac{\beta(b-1)\sqrt d}
{\sigma\nu_{b,d}^2}
\Phi_\beta(A_0).
\]
Since
\[
\Phi_\beta(A_0)\le \log\kappa(A_0),
\]
we get
\[
\E[\tau_{\kappa_\star}]
\le
C
\frac{\beta(b-1)\sqrt d}
{\sigma\nu_{b,d}^2}
\log\kappa(A_0).
\]

For fixed block size \(b\), \(\nu_{b,d}^2=\Theta_b(1)\). Thus, in the
spiky regime,
\[
\E[\tau_{\kappa_\star}]
=
O_{\kappa_\star,b,\sigma}
\left(
\beta\sqrt d\,\log\kappa(A_0)
\right).
\]
With the usual choice \(\beta=O_{\kappa_\star}(\log d)\), this becomes
\[
\E[\tau_{\kappa_\star}]
=
O_{\kappa_\star,b,\sigma}
\left(
\sqrt d\,\log d\,\log\kappa(A_0)
\right).
\]
This is a genuine \(\sqrt d\) improvement over the worst-case
\(O(d\log d\,\log\kappa(A_0))\) bound.

\subsection{Establishing intuition: First-order effect of a row correction}
\label{app:eigenvalue-correction}

We now justify the illustrative eigenvalue calculation in
Section~\ref{sec:practical}; this example is not used in the
proof of Theorem~\ref{thm:haar-block-row-min}.
Consider \(\widetilde A=I+(\lambda-1)vv^\top\), with \(\lambda>0\),
\(\lambda\neq1\), and \(\|v\|=1\). The eigenvalue \(\lambda\)
is simple, with eigenvector \(v\), while all remaining
eigenvalues equal \(1\).

For distinct coordinates \(i,j\), the congruence update
\[
    \widetilde A(\alpha)=(I+\alpha e_ie_j^\top)A(I+\alpha e_ie_j^\top)^\top.
\]
Expanding around \(\alpha=0\) gives
\[
    \widetilde A(\alpha)
    =\widetilde A+\alpha(e_ie_j^\top \widetilde A+\widetilde Ae_je_i^\top)+O(\alpha^2).
\]
By first-order eigenvalue perturbation, the eigenvalue
\(\lambda(\alpha)\) originating from \(\lambda\) satisfies
\[
\begin{aligned}
    \lambda'(0)
    &=v^\top(e_ie_j^\top \widetilde A+\widetilde Ae_je_i^\top)v\\
    &=2\lambda v^\top e_ie_j^\top v
     =2\lambda v_iv_j.
\end{aligned}
\]
Since \(\widetilde A_{ij}=(\lambda-1)v_iv_j\) for \(i\neq j\),
\[
    \lambda(\alpha)
    =\lambda+\frac{2\alpha\lambda}{\lambda-1}\widetilde A_{ij}
    +O(\alpha^2).
\]
For \(\widetilde A_{ij}<0\) and sufficiently small \(\alpha>0\),
the exceptional eigenvalue therefore moves toward \(1\),
whether \(\lambda>1\) or \(\lambda<1\).

This calculation tracks only the exceptional eigenvalue
under a single row correction. Other eigenvalues may also
change, so it does not establish a decrease in the
condition number of the full matrix.

\subsection{Empirical evidence for the unblocked load bound}
\label{app:drift-m-bound}

The proof in Section~\ref{sec:theory} uses block selection to control the load
of the row-min update.  We test whether the same type of control holds for the
unblocked selector used in the practical algorithm.  For a mixed matrix
\(A=Q\Lambda Q^\top\), define
\[
    \tau_i(A)=\arg\min_{j\neq i} A_{ij},
    \qquad
    m(A)=\|\mathcal R(A)\|_1
    =
    \max_j |\{i:\tau_i(A)=j\}|.
\]
Thus \(m(A)\) is the maximum number of rows whose minimum off-diagonal entry
selects the same column.  The step-size scaling used in our simulations is
supported if
\[
    \frac{S_\beta(A)}{\sqrt{m(A)}}=\Omega(1),
\]
where \(S_\beta\) is the spectral drift used in
Lemma~\ref{lemma:maintext_drift} and defined in~\eqref{eq:Sbeta}.  We therefore
measure this ratio directly after Haar mixing with global argmin selection, with
no blocking.

Throughout this experiment we set \(\beta=\log d\).  We generate centered
log-spectra \(x\in\mathbb R^d\), with \(\lambda_i=\exp(x_i)\), and retain only
spectra with \(\Phi_\beta(\Lambda)\ge 1\).  To search for possible failures of
the bound, we sweep several adversarial spectral families: top spikes
(\(k\) large eigenvalues), bottom holes (\(k\) small eigenvalues), two-block
spectra (a fraction \(f\) at \(+h\) and the rest at \(-h\)), power-law spectra
(\(x_i=h\,\mathrm{sign}(t_i)|t_i|^p\)), and lognormal spectra
(\(x_i=\sigma g_i\), \(g_i\sim N(0,1)\)).  Concretely, we vary
\[
    k\in\{1,2,4,\lfloor\sqrt d\rfloor,d/8,d/2\},\quad
    f\in\{0.05,0.10,0.25,0.50,0.75,0.90\},
\]
\[
    h\in\{1,\log d,2\log d,\sqrt d\},\quad
    p\in\{0.25,0.5,1,2,4\},\quad
    \sigma\in\{0.5,1,2,4\}.
\]
These families target concentrated outliers, low-eigenvalue holes, balanced
two-cluster spectra that may create collisions, smooth spectral decay, and
irregular heavy-tailed spectra.

\begin{figure}[t]
    \centering
    \includegraphics[width=0.48\linewidth]{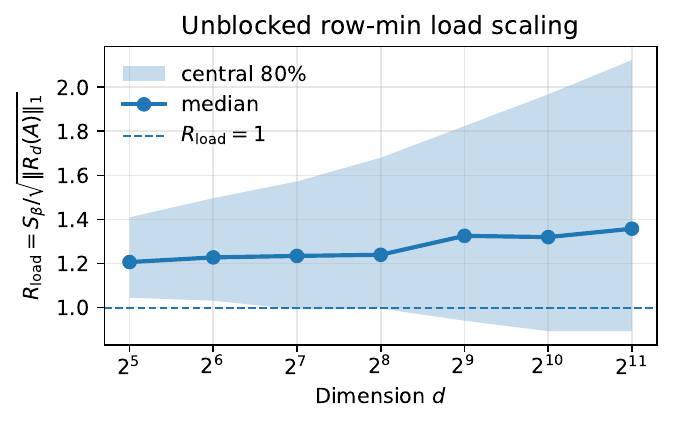}
    \hfill
    \includegraphics[width=0.48\linewidth]{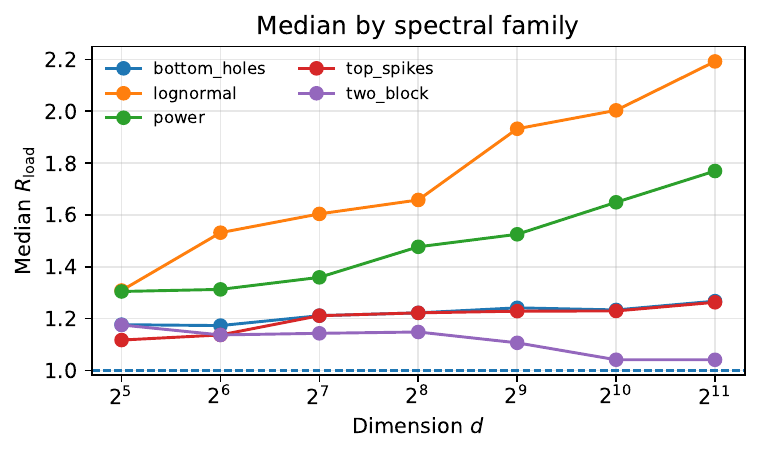}
    \caption{\textbf{Empirical load scaling for the unblocked selector.}
    \textit{Left:} distribution of \(S_\beta/\sqrt{m(A)}\) versus dimension,
    aggregated over all retained spectra and Haar trials.
    \textit{Right:} median \(S_\beta/\sqrt{m(A)}\) by spectral family.
    In all tested cases the ratio remains bounded away from zero.}
    \label{fig:unblocked-load-bound}
\end{figure}

For each retained spectrum and dimension, we draw \(256\) independent Haar
matrices \(Q\), form \(A=Q\Lambda Q^\top\), compute the exact global row-min
selector \(\tau_i(A)\), and record \(m(A)\).  We estimate
\(S_\beta(\Lambda)\) independently using \(3000\)
Dirichlet\((1/2,\ldots,1/2)\) samples, matching the squared coordinates of a
Haar row. Figure~\ref{fig:unblocked-load-bound} summarizes the
results, and Table~\ref{tab:unblocked-load-summary} reports the main aggregate
statistics by dimension.

Across the tested dimensions and spectral families,
\(S_\beta(\Lambda)/\sqrt{m(A)}\) remains bounded away from zero.  The smallest
values occur for the most symmetric high-collision cases, especially half-mass
two-block spectra and the matching top-spike and bottom-hole constructions.
Even there, the ratio stays well above zero, supporting the empirical relation
\[
    \frac{S_\beta(\Lambda)}{\sqrt{m(A)}}=\Omega(1).
\]
Substituting this into the one-step drift bound gives
\[
    \alpha_*'
    =
    \Theta\!\left(\frac{1}{\sqrt{\beta\,m(A)\,d}}\right),
\]
matching the blocked rate up to replacing the block load by the actual
unblocked load.

\begin{table}[t]
\centering
\small
\begin{tabular}{rccccccc}
\toprule
\(d\) & \# spectra
& min \( \frac{S_\beta}{\sqrt m} \)
& 10th perc. \( \frac{S_\beta}{\sqrt m} \)
& median \( \frac{S_\beta}{\sqrt m} \)
& median \(m\) & 90th perc. \(m\) & max \(m\) \\
\midrule
32   & 67 & 0.814 & 1.045 & 1.207 & 5 & 15  & 24   \\
64   & 67 & 0.804 & 1.031 & 1.228 & 7 & 22  & 42   \\
128  & 75 & 0.810 & 0.993 & 1.235 & 8 & 37  & 81   \\
256  & 75 & 0.753 & 0.996 & 1.240 & 9 & 117 & 152  \\
512  & 76 & 0.706 & 0.941 & 1.326 & 9 & 121 & 289  \\
1024 & 76 & 0.706 & 0.893 & 1.320 & 9 & 265 & 561  \\
2048 & 76 & 0.707 & 0.894 & 1.358 & 9 & 430 & 1085 \\
\bottomrule
\end{tabular}\medskip
\caption{\textbf{Aggregate statistics for the unblocked load experiment.}
Each row aggregates over all retained spectra and Haar trials at the given
dimension.}
\label{tab:unblocked-load-summary}
\end{table}

\section{Reproducibility details}
\label{appx:reproducibility}

\subsection{Compute resources}
\label{appx:compute-resources}

All experiments were run on a machine with an Intel Xeon Gold 6438M CPU,
125GiB system memory, and four NVIDIA RTX 6000 Ada Generation GPUs.  Reported experiments used a single
GPU.  The software stack used Python 3.12.4, PyTorch 2.4.0+cu121, CUDA 12.8
tooling, and Triton 3.0.0.

The transformer diagnostics used \(9\) depths and \(10\) independent runs per
depth, for \(90\) total training runs requiring approximately \(90\) cumulative
GPU-hours.  The RAPID numerical experiments were run on the same hardware using
PyTorch/Triton kernels, explicit CUDA synchronization for timing, and GPU fast
Walsh--Hadamard transforms \citep{dao2024fast}.

\subsection{Existing assets and licenses}
\label{appx:assets-licenses}

The experiments use synthetic matrices generated by our scripts and public
SuiteSparse Matrix Collection instances.  SuiteSparse matrices were accessed
through \texttt{ssgetpy}; the SuiteSparse Matrix Collection is credited in the
main text, and \texttt{ssgetpy} is distributed under the MIT/X11 license.

Our implementation uses Python 3.12.4, PyTorch 2.4.0+cu121, Triton 3.0.0,
CUDA tooling, NumPy, SciPy, and \texttt{fast-hadamard-transform}.  PyTorch is
distributed under a BSD-style license, Triton under the MIT license, NumPy and
SciPy under BSD-style licenses, \texttt{fast-hadamard-transform} under the
BSD-3-Clause license, and the CUDA Toolkit is governed by the NVIDIA CUDA
Toolkit EULA.  We use these packages only as software dependencies and do not
redistribute modified versions of them.